\documentclass{article}

\usepackage[margin=1.05 in]{geometry}

\usepackage{ulem}

\usepackage[round]{natbib}
\PassOptionsToPackage{numbers, sort&compress}{natbib}

\usepackage[dvipsnames]{xcolor}

\usepackage{graphicx}
\usepackage{wrapfig}
\usepackage{subcaption}
\usepackage{xfrac}
\usepackage{old-arrows}
\usepackage{array}

\usepackage{wrapfig}

\usepackage{empheq}

\usepackage{commath}
\usepackage{mathtools}
\usepackage{bm}
\mathtoolsset{showonlyrefs}
\usepackage{algorithm}
\usepackage{algorithmic}
\usepackage[dvipsnames]{xcolor}

\usepackage{enumitem}
\usepackage[utf8]{inputenc} 
\usepackage[T1]{fontenc}    
\usepackage{hyperref}       
\usepackage{url}            
\usepackage{booktabs}       
\usepackage{amsfonts}       
\usepackage{nicefrac}       

\usepackage{microtype}      
\usepackage{xcolor}         
\usepackage{xfrac}
\usepackage{wrapfig}
\usepackage{enumitem}
\usepackage{amsmath}
\usepackage{amssymb}
\usepackage{enumitem}

\newlist{assumptions}{enumerate}{1}
\setlist[assumptions]{label=(A\arabic*), ref=A\arabic*} 

\usepackage{hyperref}
 \hypersetup{
     colorlinks=true,
     linkcolor=blue,
     filecolor=blue,
     citecolor = blue,      
     urlcolor=magenta,
 }

\usepackage{amssymb}

\usepackage{amsthm}
\usepackage{amsmath}
\usepackage{mathtools}
\mathtoolsset{showonlyrefs}
\usepackage{dsfont}
\usepackage{amssymb}
\usepackage{latexsym}
\usepackage{verbatim}
\usepackage{xfrac}
\usepackage{physics}
\usepackage{enumitem}

\usepackage{bbm}

\usepackage{amsthm}

{\begin{list}               
    {$\bullet$ \hfill}{
        \setlength{\leftmargin}{\parindent}
        \setlength{\parsep}{0.04\baselineskip}
        \setlength{\itemsep}{0.5\parsep}
        \setlength{\labelwidth}{\leftmargin}
        \setlength{\labelsep}{0em}}
    }
{\end{list}}

\providecommand{\cref}[1]{Chapter~\ref{chap:#1}}

\providecommand{\R}{\ensuremath{\mathbb{R}}}

\providecommand{\inprod}[1]{\left\langle#1\right\rangle}

\newcommand{\E}{\mathbb{E}}
\renewcommand{\P}[1]{\mathbb{P}\left[#1\right]}

\newcommand{\Ea}[1]{\E\left[#1\right]}

\newcommand{\eg}{\varepsilon_g}
\newcommand{\ploc}{p^\circ}
\newcommand{\pglob}{p^\star}

\DeclareMathOperator{\arccosop}{arccos}

\DeclareMathOperator*{\sgn}{sign}

\theoremstyle{plain}

\theoremstyle{plain}
\newtheorem{theorem}{Theorem}[section]

\theoremstyle{plain}

\theoremstyle{plain}
\newtheorem{remark}[theorem]{Remark}
\theoremstyle{plain}
\newtheorem{conjecture}[theorem]{Conjecture}
\theoremstyle{plain}
\newtheorem{result}[theorem]{Result}

\usepackage{authblk}

\title{Towards optimal selection schedules for online active learning}

\author[1]{Hugo Cui}
\author[2,3]{Francesca Mignacco}
\author[4]{Francesco Mori}

\affil[1]{\small  Université Paris-Saclay, CNRS, Laboratoire de mathématiques d’Orsay, 91405, Orsay, France
}
\affil[2]{\small  Joseph Henry Laboratories of Physics and Lewis–Sigler Institute, Princeton University, Princeton NJ 08544, USA
}
\affil[3]{\small  Initiative for the Theoretical Sciences, The Graduate Center,
City University of New York, 365 Fifth Ave, New York NY 10016, USA 
}
\affil[4]{\small Center of Mathematical Sciences and Applications, Harvard University, 20 Garden Street, Cambridge MA 02138, USA
}

\begin{document}

\date{}

\maketitle

\begin{abstract}
We consider the problem of online linear classification, where a stream of unlabeled samples is available, and annotation is costly. We investigate the optimal query policy, that prescribes whether and when a sample should be labeled and used for a training step, or instead discarded, so as to minimize a given objective. We focus on the realizable case, under Gaussian design, in the high-dimensional regime where data dimension and number of unlabeled samples grow proportionally. We consider two objectives: 1) the generalization error under a labeling budget, and 2) a cumulative cost function where mistakes have unit cost and queries cost $\Delta$. We first show that the myopic policies that greedily minimize at all times the increment in the objective correspond to either small or intermediate-margin policies, that prescribe the selection of samples whose margins fall within certain intervals that shrink with time. Within this class, we numerically ascertain the optimal schedules for the selection thresholds that minimize the final values of the objectives. Taking inspiration from the myopic policy for the cumulative cost, we design a very simple query policy schedule that we compare to other proposed heuristics, on synthetic data and real medical image datasets. Finally, we discuss myopic policies for two-layer neural networks, and exemplify how naive margin-based policies can cease to be optimal.
\end{abstract}

\section*{Introduction}

Many breakthroughs in modern deep learning were achieved by training models on ever-accruing volumes of data. Yet, there exist important domains of application where labeled data is scarce. Examples include medical diagnosis \citep{liu2004active}, rare languages \citep{kholodna2024llms} or molecular modeling \citep{felis2025falcon}, where data annotation requires either a human expert, or expensive ab-initio simulations, resulting in prohibitive labeling costs. In such scenarios, it is thus essential to \textit{carefully select} which samples, among a stream of unlabeled data, should be labeled. 

In the present manuscript, we consider the online setting, where unlabeled data points are viewed sequentially. For each sample, the model can decide either to output a prediction, and incur a cost in case of mistake, or to query the label from a (e.g., human) expert, paying the associated annotation cost, and use the thus labeled instance to perform a gradient step, furthering its training. For example, in medical applications, a diagnosis must be issued by the model to every patient, unless it is determined further examinations need to be performed first by an expert. Querying such additional examinations is however typically costly, and thus must be done sparingly. If performed however, such examinations provide an additional data point that can improve the model accuracy, thereby reducing the frequency of future expert queries. How then to choose which labels to query, and when? 

Many selection policies were proposed in the literature on online active
learning \citep{dasgupta2005analysis, hanneke2014theory} and selective sampling
\citep{reveillard2025minimizing, hanneke2021toward, cesa2004worst,
cesa2009robust}, some of which were shown to achieve the optimal \textit{rates}
of mistakes, queries or accuracy. These works, however, typically fix the form
of the query policy in advance and analyze its performance. Here, we take a
complementary approach and ask what the \textit{optimal} policy schedule is. We consider the classical problem of online binary linear
classification in the realizable case with the Adatron learning rule
\citep{anlauf1989adatron, motzkin1954relaxation}, under Gaussian design, in the
high-dimensional limit where the dimension $d$ of the samples is comparable to
the total number $n$ of unlabeled data \citep{seung1992query, kinzel1990improving,
watkin1992selecting, engel2001statistical, rattray1997globally}. Using techniques from statistical physics and control theory, we
precisely ascertain the \textit{optimal query policy schedule} among a class of
margin-based policies that minimizes, down to the constant, a given cost
objective --- either the generalization error under a labeling budget, or the
cumulative number of mistakes and queries.

More precisely, we distinguish between \textit{myopic policies}, which greedily minimize at all times
the instantaneous increment in the objective, and \textit{horizon-optimized
schedules}, obtained by numerically optimizing the selection policies across all
training times to minimize the objective at the end of training. Two classes of margin-based policies will prove of particular interest: \textit{inner-margin}
policies, which query only samples within a threshold distance of the decision
boundary, and \textit{intermediate-margin} policies, which query samples whose
distance lies between two thresholds, thus discarding both the closest and the
farthest samples. Our \textbf{main contributions} are:\looseness=-1
\begin{itemize}
    \item We first study policies that minimize the generalization error given $n$ unlabeled samples, under a total query budget $\gamma n$, with $\gamma\in(0,1]$. We show that the myopic strategy is of the intermediate-margin type.
    Within this class, we then compute the horizon-optimized threshold schedules, which achieve a lower final generalization error than random, myopic, and inner-margin strategies.
    \item We then consider the setting where mistakes have unit cost and queries cost $\Delta>0$.  We derive an exact expression for the myopic policy, showing that it is an inner-margin rule with shrinking threshold. We propose a simple selection heuristic inspired therefrom, and compare its performance with other baselines on synthetic data and real medical image datasets. In the same setting, the horizon-optimized schedule is again an inner-margin rule, with higher query rate early in training, outperforming both the myopic and the random policies, with gains that grow with $\Delta$.
    \item Finally, we study myopic policies for cumulative-cost minimization in \textit{two-layer neural networks}, allowing them to take as an argument the full vector of preactivations. We exemplify how inner-margin policies naively applied to the network readout are suboptimal.
\end{itemize}

The code used in the manuscript can be found in the following \href{https://github.com/francescomori/optimal_selection}{repository}.

\subsection*{Related works}

\paragraph{Stream-based active learning.} A now mature body of works has studied the \textit{label complexity} of online active learning, ascertaining the minimal number of queries necessary to achieve a given accuracy \citep{freund1997decision,freund1997selective,hanneke2014theory}. The realizable case where the two classes lie in distinct half-spaces, in particular, has been the object of refined analyses. \cite{dasgupta2005analysis, yan2017revisiting} construct active learning algorithms achieving the optimal $\Tilde{O}(d\log \sfrac{1}{\epsilon})$ label complexity required to reach accuracy $1-\epsilon$. The present work complements these results in the high-dimensional Gaussian setting, and investigates, given a number $n$ of unlabeled instances and a query budget $\gamma n$, the minimal generalization error achievable, down to the constant, within a class of margin-based query policies. \looseness=-1

\paragraph{Selective sampling.} In problems where the learner needs to output a prediction after viewing each sample, one aims at minimizing the \textit{cumulative number of mistakes} over training, rather than the final accuracy \citep{reveillard2025minimizing, hanneke2021toward, gangrade2021online, yu2026distribution,goel2023adversarial}. Formally, a selective sampling algorithm must determine at each iteration whether to issue a prediction, or instead query the label, with the goal of minimizing the overall number of mistakes and queries. The latter can be controlled for a variety of algorithms, as studied in \cite{reveillard2025minimizing, hanneke2021toward}. Some of these guarantees however become vacuous in high-dimensional settings. In the present work, in the case where mistakes have unit cost and queries have cost $\Delta$, we precisely characterize the myopic policy for the cumulative cost, and ascertain its large-time asymptotic behavior down to the constant. 

\paragraph{Optimal schedules for high-dimensional online learning.} In contrast to most works on online active or selective learning, which offer theoretical guarantees for a specific algorithm, this paper investigates the \textit{optimal} query strategy. Formally, this amounts to ascertaining the optimal schedule of selection policies, such that the learning process driven thereby realizes minimal cost or error, while respecting a query budget. The large dimensionality $d\gg 1$ of the space in which the process is embedded, however, seemingly poses significant barriers to the application of standard control-theoretic approaches. For Gaussian data nevertheless, the training dynamics (including instantaneous accuracy and mistake rates) can be succinctly described by a finite set of \textit{scalar} ordinary differential equations (ODEs) \citep{saad1995exact, saad1995line, biehl1995learning, goldt2019dynamics, arous2021online, collins2024hitting}, on which standard variational and control analyses can be applied. This route allows for a precise characterization of optimal hyperparameter schedules, including learning rate \citep{rattray1997globally, rattray1998analysis, bordelon2026theory}, task ordering \citep{mori2025optimal} or dropout \citep{mori2025analytic,mignacco2025statistical}. A generalization to \textit{functional} hyperparameter schedules was exposed in \cite{rattray1997globally}. We draw from this conceptual framework to find optimal selection policies for active half-space learning in high dimensions.

\paragraph{Active learning half-spaces in high dimensions.} The question of active half-space learning in the proportional high-dimensional regime $n\asymp d$ has received a long and sustained attention, notably in the pool-based setting \citep{cui2021large, kolossov2023towards, dohmatob2025less, sorscher2022beyond,askari2025improving, cui2026asymptotic}, focusing on inner-margin policies. \cite{derenyi1994generalization} study the converse problem where samples farthest from the ground-truth separator are retained, providing a characterization of the optimal threshold.
The online setting was first studied by \cite{seung1992statistical, kinzel1990improving, watkin1992selecting, zhou2019active}, in the case where the learner is allowed to design and synthesize specific samples at each step and query the accompanying label. \cite{ kinzel1990improving, watkin1992selecting} in particular study the prescription of designing samples orthogonal to the running weights, which corresponds to an inner-margin policy with vanishing threshold. \cite{parrondo1993error} conversely consider the case where large-margin samples are retained, while \cite{jung1996selection} study gradient-based selections. Closer to our work, \cite{mai2022effectiveness} derive a sharp characterization of online stream-based active learning with repeated retrainings, and determine the optimal batch size for a two-step procedure. All the above works, however, assume a specific functional form for the selection policy. Here, we derive myopic policies analytically and numerically compute horizon-optimized candidates within the same functional class. While in certain settings these policies indeed correspond to inner-margin policies, as studied in prior works, we find that they can also prescribe the retention of samples at \textit{intermediate} margins. The present manuscript also offers a precise characterization of the optimal \textit{schedule} of such selection functions over the course of training.\looseness=-1

\section{Problem setup}
\label{sec:setup}

\paragraph{Online active classification.} We consider the problem of active online learning with a linear classifier, under Gaussian design. A stream of $n$ unlabeled samples $(x_i)_{i\in [n]}$, drawn i.i.d from a $d-$dimensional Gaussian distribution $\mathcal{N}(0_d, \sfrac{I_d}{d})$, is sequentially observed by the learner. At each round $i\in [n]$, the learner can 
\begin{itemize}
    \item Query an expert to annotate the sample, thereby obtaining the associated label $y_i$. We consider the realizable setting, where there exists an (unknown) ground truth vector $\beta \in\R^d$ with $\lVert \beta\rVert^2=d$ that perfectly separates the data, namely $y_j=\mathrm{sign}(\inprod{\beta, x_j})$ for all $j \in[n]$. With the full covariate-label pair $(x_i, y_i)$ now available, the learner may then perform a gradient update on the weights $w_{i}$ of a linear classifier
    \begin{align}
    \label{eq:discrete_GD_step}
        w_{i+1}=w_i-\eta \ell^\prime(\inprod{w_i,x_i}y_i)y_i x_i,
    \end{align}
    with learning rate $\eta>0$ and loss $\ell:\R\to \R_+$. Note that this training process corresponds to stochastic gradient descent (SGD) at batch size $1$.\looseness=-1
    \item Alternatively, the learner may choose to not pay the cost of querying an expert, and instead issue a prediction $\sgn(\inprod{w_i, x_i})$ for the unknown label $y_i$ of sample $x_i$, at the risk of making a mistake whenever $\sgn(\inprod{w_i, x_i}) \ne y_i$.
\end{itemize}
 In line with prior works on active binary classification \citep{engel2001statistical, dasgupta2005analysis, cesa2009robust, sculley2007online}, we focus on the $0-$threshold squared hinge loss $ \ell(z)=\sfrac12\min(z,0)^2$, for which the gradient update \eqref{eq:discrete_GD_step}  reduces to the celebrated \textit{Adatron} rule \citep{anlauf1989adatron, motzkin1954relaxation}. The latter has been repeatedly leveraged in active learning analyses for its favorable convergence properties: in the passive case, the Adatron rule achieves a fast $\sfrac{1}{i}$ decay in generalization error \citep{engel2001statistical} while in the active setting, it can lead to exponential decay with the number of labeled instances \citep{dasgupta2005analysis}. For extensions to the logistic loss, see App.~\ref{app:other_losses}.

\paragraph{Selection policy.} At every round $i\in[n]$, the choice of the learner can be encoded by a binary variable $\sigma_i\in\{0,1\}$, with the learner choosing to query the label (resp. issue a prediction) if and only if $\sigma_i=1$ (resp. $\sigma_i=0$). In the case of linear classifiers, it is natural to base this choice on the margin (output) $\inprod{w_i,x_i}$ of the model for the running sample $x_i$. We thus model $\sigma_i$ as a Bernoulli variable of law $\mathcal{B}(p_i(\inprod{w_i, x_i}))$, where the sequence $(p_i)_{i\in[n]}$ of functions $p_i:\R\to [0,1]$ denotes the schedule of \textit{selection policies} according to which the learner makes decisions. Note that we thus allow the policy to vary in time, and to be stochastic. Interestingly, the optimal policies derived below will correspond to \textit{deterministic} selections, where $p_i$ takes values in $\{0,1\}$. A particularly well-known instance of such deterministic rules are hard inner-margin rules $p_i(\cdot)=\mathds{1}[|\cdot|\le \kappa]$ \citep{tong2001support, roth2006margin}, which prescribe to label samples whenever they fall within a certain distance $\kappa$ of the decision boundary, with the intuition that the closest samples are the hardest to classify for the current model, hence the most informative. Many other margin-based policies tailored to online Adatron or perceptron learning were proposed in the literature, see for instance \cite{dasgupta2005analysis, sculley2007online, cesa2004worst}, and \cite{cacciarelli2024active} for a survey.


\paragraph{Cost metrics.} The present manuscript aims at ascertaining the \textit{optimal} policy schedule $p_i(\cdot)$. There exist two natural notions of cost with respect to which policies may be compared. The first is the \textit{final} generalization error, defined as the misclassification error
\begin{align}\label{eq:gen_err_definition}
    \varepsilon_{g,i}=\P{\inprod{w_i, x}y<0},
\end{align}
at the final round $i=n$.
In the above display, the probability bears over test samples $x,y$ from the same distribution as the stream of training samples. We investigate policies minimizing $\epsilon_{g,n}$ with a constraint on the total number $\mathcal{Q}_n$ of queried samples:
\begin{equation}\label{eq:Qk}
    \mathcal{Q}_n\equiv \sum\limits_{i=1}^n \sigma_i= \lfloor\gamma \,n\rfloor,
\end{equation}
where $\gamma\in(0,1]$ is the labeling budget.
The final generalization error is the metric most often adopted in the active learning literature, and describes situations where the model is audited only post-training, when solely its final performance matters. To give an analogy in human learning, the same setup describes the situation of a student, whose performance at exercise sessions is inconsequential, provided they perform well at the final exam. 

Selective sampling is, instead, concerned with controlling the \textit{cumulative} number of errors and queries over training. If mistakes have unit cost while queries have relative cost $\Delta $, the cumulative cost at round $k$ is 
\begin{align} \label{eq:Rk}
    \mathcal{R}_k=\frac{1}{d}\left(\mathcal{M}_k+\Delta ~\mathcal{Q}_k\right)\,.
\end{align}
The cumulative number of queries $\mathcal{Q}_k$ is defined in \eqref{eq:Qk} and the cumulative number of mistakes is given by
\begin{align}\label{eq:Mk}
    \mathcal{M}_k=\sum\limits_{i=1}^k (1-\sigma_i)\mathds{1}[\inprod{w_i, x_i}y_i<0]\,. 
\end{align}
For instance, in the case of medical diagnostics, a diagnosis needs to be presented to \textit{each} patient, posing the challenge to properly balance the costs of a wrong diagnosis and that of asking for expensive human expertise. \looseness=-1

\begin{figure}
\centering
\includegraphics[width=0.4\linewidth]{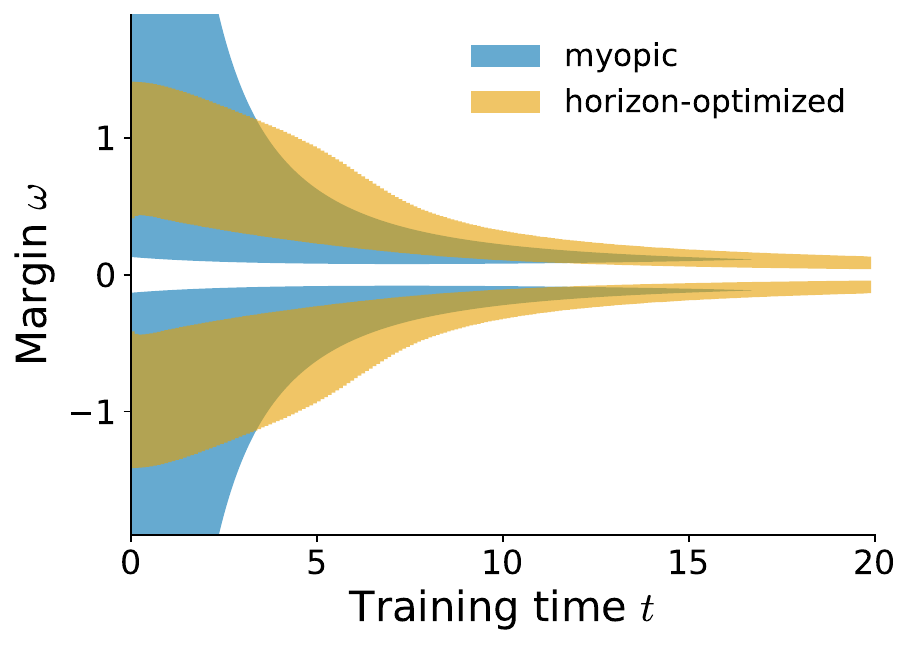}
  \hspace{.8cm}
\includegraphics[width=0.49\linewidth]{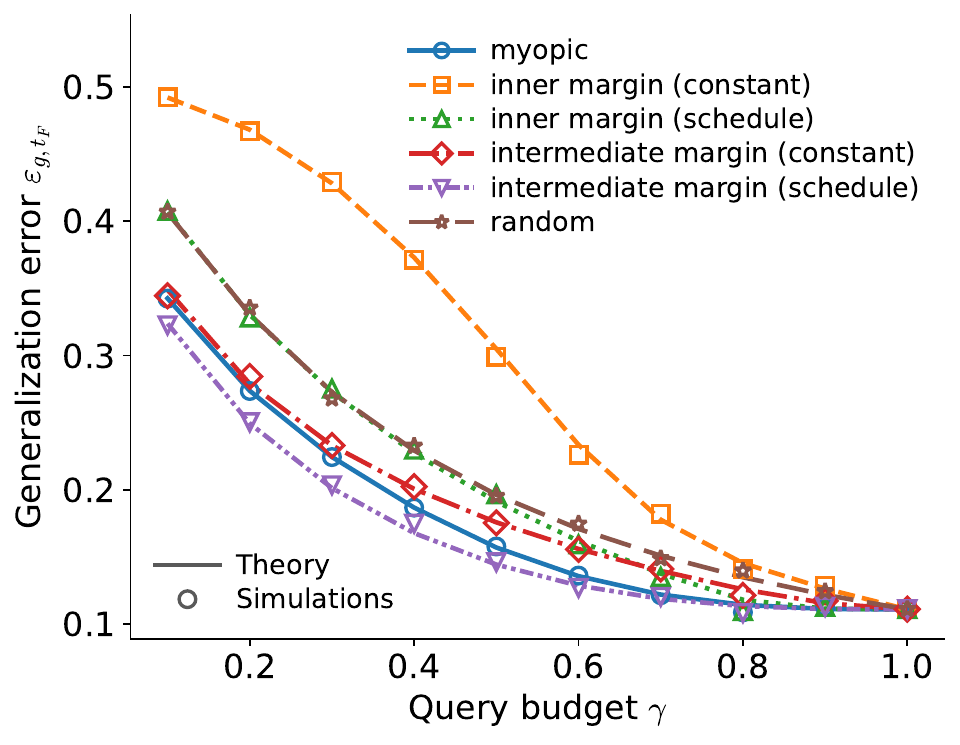}

  \caption{(\textbf{left}) Selection policies in the $(t,\omega)$ plane for the final-error minimization. Colored regions indicate
  $p_t(\cdot)=1$. The myopic rule $\ploc_{\epsilon,t}(\cdot)$ \eqref{eq:ploc_eg} is shown in blue. The horizon-optimized intermediate-margin schedule $p^{\star}_{\epsilon,t}(\cdot)=\mathds{1}[a_t^\star\leq|\cdot| \leq b_t^\star]$ \eqref{eq:pglob_eg}, with $a_t^\star$ and $b_t^\star$ set to minimize the final error, is shown in yellow. Parameters: $\gamma=0.4$, $\eta=0.5$,
    $m_0=0$, $q_0=0.15$, $t_F=20$. (\textbf{right}) Final generalization error $\varepsilon_{g,t_F}$ vs.~query budget
    $\gamma$. Parameters: $\eta=0.5$, $m_0=0$, $q_0=0.15$, $t_F=20$. Lines represent theoretical predictions while symbols mark single-trajectory SGD simulations at $d = 1000$. For details on the policies see Table \ref{tab:strategies}.}
  \label{fig:sel_epsilon}
\end{figure}

\paragraph{Myopic and horizon-optimized rules.} Our analysis thus sets out to describe the \textit{horizon-optimized} policies $\pglob_{\epsilon,i}$, $\pglob_{\mathcal{R},i}$, respectively minimizing the average error $\Ea{\epsilon_{g,n}}$ under a labeling budget $\gamma\in (0,1]$ or the average total cumulative cost $\Ea{\mathcal{R}_n}$. From prior works in online learning in the high-dimensional regime $n\asymp d$ \citep{saad1995exact,saad1995line, goldt2019dynamics}, one expects that for a given policy the random variables $\epsilon_{g,n}$ and $\mathcal{R}_n$ concentrate around their expectations as $n,d\to \infty$. Heuristically, policies resulting in small costs in expectation are thus also anticipated to display good performance on individual trials, as we illustrate in Figs.~\ref{fig:sel_epsilon} (right) and \ref{fig:sel_R} (right). 

We also consider \textit{myopic} policies, which at each round minimize the expected change of the objective, conditional on the model weights at the beginning of the round. Formally, the myopic policy $\ploc_{\epsilon,i}$
(resp.\ $\ploc_{\mathcal{R},i}$) minimizes
$\Ea{\epsilon_{g,i+1}-\epsilon_{g,i} }$
(resp.\ $\Ea{\mathcal{R}_{i+1}-\mathcal{R}_{i}}$),
where the expectation bears on the sample $(x_i,y_i)$ and the query decision
$\sigma_i$. Because they ignore the
effect of the current choice on future rounds, myopic policies can prove
suboptimal whenever queries that seem costly early in training pay off later on. For instance, for $\Delta>1/2$ a myopic learner never queries, since random-chance error is $1/2<\Delta$. However, a strategy that minimizes the final cost $\mathcal{R}_n$ could in principle query even at $\Delta>\sfrac{1}{2}$. For simplicity, we focus on $\Delta<1/2$ throughout.\looseness=-1


\section{Closed-form description of the training dynamics}

In the high-dimensional regime $d\gg 1$, identifying optimal selection strategies is hampered by the curse of dimensionality. We circumvent this challenge by reducing the high-dimensional SGD dynamics in Eq.~\eqref{eq:discrete_GD_step} to scalar ODEs bearing on a small set of summary statistics, also known as \textit{order parameters} in statistical physics \citep{saad1995exact}. Following \cite{mignacco2025statistical}, we then apply standard control-theoretic methods to this low-dimensional description to find the optimal selection schedules.

\begin{result}[Exact ODEs for the learning dynamics] \label{res:ODEs}
Let $w_i$ denote the weights of the learner after $i$ steps of the online SGD dynamics in \eqref{eq:discrete_GD_step}. In the limit $d,i\to\infty$ with $t=\sfrac{i}{d}= \mathcal{O}_d(1)$, the scalar summary statistics
\begin{equation}\label{eq:OPs_definition}
    m_t=\frac{\langle w_{t d},\beta\rangle}{d}\,,\quad \quad q_t=\frac{\langle w_{t d},w_{t d}\rangle}{d}
\end{equation}
concentrate in probability to the solution at time $t$ of the following deterministic ODEs
\begin{align}\label{eq:ODEs}
  \frac{\dd m_t}{\dd t}
    = f_m (m_t,q_t;p_t)\,,&&
  \frac{\dd q_t}{\dd t}
    = f_q (m_t,q_t;p_t)\,,
\end{align}
where $f_m$ and $f_q$ are functions of $m_t$ and $q_t$ and functionals of the selection policy $p_t$. The expressions of $f_m$ and $f_q$ are given in App.~\ref{app:ODEs}. We refer to $t$ as the training time and $t_F=\sfrac{n}{d}$ as the time horizon. 
\end{result}
Fig.~\ref{fig:concentration} in App.~\ref{app:ODEs} shows that this theoretical prediction matches a single instance of numerical simulations already at $d=1000$. The ODEs capture the time evolution of the squared norm $q_t$ of the weights and their overlap $m_t$ with the ground truth vector $\beta$. In the remainder of the paper, we work within this continuous-time description and, with a slight abuse of notation, index all quantities introduced in Sec.~\ref{sec:setup} by $t$, e.g.\ $\mathcal{R}_t = \mathcal{R}_{td}$.

\begin{remark}
    The cost metrics in Eq.~\eqref{eq:gen_err_definition} and \eqref{eq:Rk} can be expressed as explicit functions of the summary statistics $m_t$ and $q_t$. The generalization error is $
\varepsilon_{g,t} = \sfrac1\pi\arccosop\!\left(\sfrac{m_t}{\sqrt q_t}\right)\,.$
Similarly, the cumulative cost $\mathcal{R}_t$ \eqref{eq:Rk} can be expressed as a function of the summary statistics $m_{t'}$ and $q_{t'}$ for $ t'\in [0,t]$, see App.~\ref{app:ODEs}.
\end{remark}

Since the cost metrics are determined by the summary statistics, which evolve
according to the low-dimensional deterministic dynamics in
Eq.~\eqref{eq:ODEs}, we can use control-theoretic methods to find optimal
training strategies. When considering inner or intermediate-margin policies, we parametrize the selection thresholds as piecewise
constant in time and optimize them by direct single shooting, using exact
gradients obtained by automatic differentiation through the discretized ODEs
in \texttt{JAX} (see App.~\ref{app:control}). The resulting problem is
non-convex, so the schedules we obtain are not guaranteed to be global optima. Nevertheless, they consistently outperform all the benchmarks we consider.\looseness=-1


\section{Minimizing the final error}
\label{sec:budget}

We first look for selection policies $p_t$ that minimize the final generalization  error $\varepsilon_{g,t_F}$ over a given time-horizon $t_F$, subject to the following continuous-time constraint on the total number of queries (cf.~Eq.~\eqref{eq:Qk})
\begin{equation}
  \frac{1}{t_F}\int_0^{t_F}\!\E[p_t(\omega_t)]\,\dd t = \gamma\,,
  \label{eq:budget}
\end{equation}
where $\E[\cdot]$ is the expected value w.r.t.~$\omega_t\sim \mathcal{N}(0,q_t)$.

\subsection{Myopic policy \texorpdfstring{$\ploc_{\varepsilon,t}$}{}}

As a first step, we investigate the myopic policy that greedily maximizes at each step the decrease in generalization error $-\sfrac{\dd\varepsilon_{g,t}} {\dd t}$. This analysis will serve as a basis for designing the horizon-optimized policy that minimizes the terminal error $\varepsilon_{g,t_F}$. Note however that such myopic policy would not satisfy the global constraint in Eq.~\eqref{eq:budget}. As a consequence, we introduce a modified myopic policy that depends on a single Lagrange multiplier $\nu$, enforcing \eqref{eq:budget} (see App.~\ref{app:control}).

\begin{result}[Myopic policy under query budget]
\label{res:loc_budget}
    The myopic policy is given by 
    \begin{equation}
        \ploc_{\varepsilon,t}(\cdot)=\mathds{1}\left[S_t(\cdot)+\nu<0\right]\,,
    \end{equation}
where $S_t(\cdot)$ is given in Eq.~\eqref{eq:slope} of App.~\ref{app:control} and $\nu>0$ is chosen to enforce Eq.~\eqref{eq:budget}.
For a wide range of parameters this results in the intermediate-margin policy
\begin{equation} \label{eq:ploc_eg}
        \ploc_{\varepsilon,t}(\cdot)=\mathds{1}[a_t\le|\cdot| \le b_t]\,,
    \end{equation}
    where the threshold values $a_t$ and $b_t$ are the positive roots of $S_t(\omega)+\nu=0$  (see Fig.~\ref{fig:grid}). 
\end{result}
Interestingly, while inner-margin selection policies are most commonly studied in the literature, Res.\,\ref{res:loc_budget} shows that selecting samples with intermediate margins can be more beneficial. We can interpret this policy by noting that the weight update \eqref{eq:discrete_GD_step} from the Adatron rule is proportional to the margin $|\omega|$ for misclassified samples. Samples with small margins therefore produce updates that are too small to justify spending the query budget on them. Conversely, at large $|\omega|$ samples are rarely misclassified,
so a query there would also be uneconomical. Furthermore, although stochastic policies are allowed, the myopic policy is deterministic almost surely, because the instantaneous objective is linear in the selection probability (see App.~\ref{app:control}).\looseness=-1

\subsection{Horizon-optimized intermediate-margin policy \texorpdfstring{$\pglob_{\varepsilon,t}$}{}}
Motivated by the functional form of the myopic policy in Res.~\ref{res:loc_budget}, we use control theory to identify the policy $\pglob_{\varepsilon,t}$ minimizing the final generalization error $\varepsilon_{g, t_F}$ within the class of intermediate-margin rules (for details, see App.~\ref{app:control}). 

\begin{result}[Horizon-optimized policy under query budget]\label{res:pglob_eg}
The horizon-optimized intermediate-margin policy is
    \begin{equation} \label{eq:pglob_eg}
\pglob_{\varepsilon,t}(\cdot)
        =\mathds{1}[a^\star_t\le|\cdot| \le b^\star_t]\;,
\end{equation}    
  where the optimal thresholding schedule is given by
\begin{equation}\label{eq:globally_eps}
(a_t^\star,b_t^\star) = \operatorname{argmin}_{\{a_t,b_t\}} \varepsilon_{g, t_F}
        \,, \text{   subject to \eqref{eq:budget}\,.}
\end{equation}
The minimization in Eq.~\eqref{eq:globally_eps} bears on $0\leq a_t\leq b_t$ for $t\in[0,t_F]$, resulting in the schedule in Fig.~\ref{fig:sel_epsilon}.
\end{result}
Fig.~\ref{fig:sel_epsilon} (left) compares the myopic policy $\ploc_{\varepsilon,t}$ \eqref{eq:ploc_eg}, in blue, with the horizon-optimized intermediate-margin policy $\pglob_{\varepsilon,t}$ \eqref{eq:pglob_eg}, in yellow. Both schedules reduce the selection interval as training progresses and classification improves at smaller margins. However, the horizon-optimized schedule can exploit the full training horizon to restrict the selection interval from the beginning, whereas the myopic one starts by querying a much broader range of margins, that early in training yield substantial immediate improvements, but it exhausts the labeling budget before reaching $t_F$.

Fig.~\ref{fig:sel_epsilon} (right) shows the final generalization error as a function of
the query budget for $\ploc_{\varepsilon,t}$ and $\pglob_{\varepsilon,t}$. As
benchmarks, we include the (time and margin independent) random policy $p=\gamma$, the constant
intermediate-margin policy $p(\cdot)=\mathds{1}[a^\star\le|\cdot|\le b^\star]$,
and the inner-margin policy $p_t(\cdot)=\mathds{1}[|\cdot|\le\kappa_t^\star]$, with
either a constant threshold $\kappa^\star$ or a schedule $\kappa^\star_t$. All free
parameters are optimized to minimize $\varepsilon_{g, t_F}$. The optimal intermediate-margin schedule achieves the lowest final error, with the largest gains at small query budget $\gamma$. Even with constant thresholds, intermediate-margin selection outperforms constant and time-dependent inner-margin policies at small and moderate budgets $\gamma$. We note that this observed sub-optimality of inner-margin selections echoes the findings of \cite{hacohen2022active, hacohen2023select, kolossov2023towards, cui2026asymptotic, tifrea2023margin} in the pool-based setting.\looseness=-1

\section{Minimizing the cumulative cost}
\label{sec:integrated_error}

\begin{figure}
\centering
\includegraphics[width=0.4\linewidth]{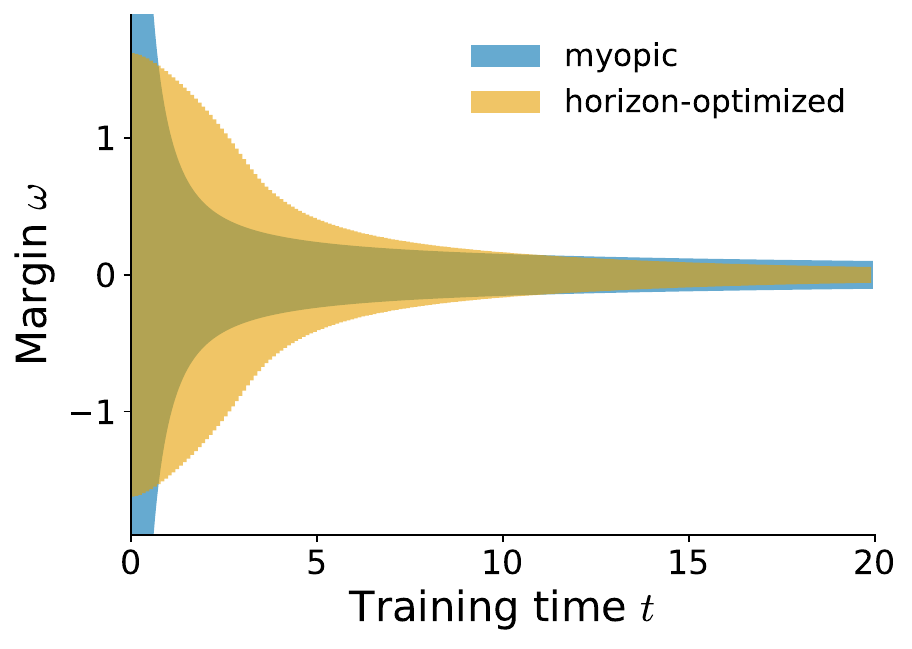}
    \hspace{.8cm}
\includegraphics[width=0.49\linewidth]{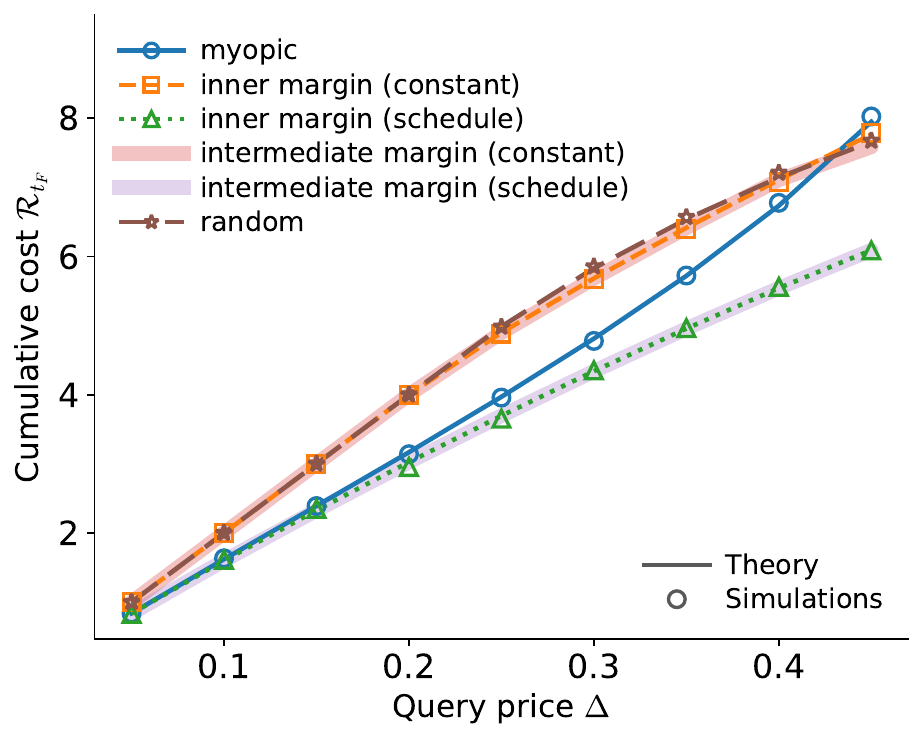}
  \caption{(\textbf{left}) Selection policies in the $(t,\omega)$ plane for cumulative-cost minimization. Colored regions indicate
  $p_t(\cdot)=1$. The myopic rule $\ploc_{\mathcal{R},t}(\cdot)$,
  greedily maximizing the cost decrease at each time, is shown in blue. The horizon-optimized inner-margin schedule $p^{\star}_{\mathcal{R},t}(\cdot)=\mathds{1}[|\cdot| \leq b_t]$, with $b_t$ set to minimize the final cumulative cost, is shown in yellow. Parameters: $\Delta=0.3$, $\eta=0.5$, $m_0=0$, $q_0=0.15$, $t_F=20$. (\textbf{right}) Cumulative cost $\mathcal{R}_{t_F}$ vs.~the query
    price $\Delta$. Parameters: $\eta=0.5$, $m_0=0$,
    $q_0=0.15$, $t_F=20$. Lines represent theoretical predictions while symbols mark single-trajectory SGD simulations at $d = 1000$. Constant and optimal intermediate-margin families collapse to the inner-margin strategies once optimized. For details on the policies see Table \ref{tab:strategies}.}
  \label{fig:sel_R}
\end{figure}
In this section, we now turn to studying policies minimizing the cumulative cost
$\mathcal{R}_{t_F}$ \eqref{eq:Rk}. We show that both the myopic policy and the horizon-optimized intermediate-margin policy are inner-margin
rules. Indeed, samples with small $|\omega|$ produce
small weight updates and are thus discarded when the objective is the final
error under a budget. However, these are also the samples the model is most
likely to misclassify. If mistakes made during training are costly, querying
them avoids a likely error, and this benefit outweighs the small size of the
resulting update.

\subsection{Myopic policy \texorpdfstring{$\ploc_{\mathcal{R},t}$}{}}

The following Result, derived in App.\,\ref{app:Localopt_R}, shows that the myopic policy $\ploc_{\mathcal{R},t}$ minimizing the instantaneous increase $\sfrac{\dd}{\dd t} \mathcal{R}_t$ in cost \eqref{eq:Rk} corresponds to a deterministic inner-margin policy \citep{tong2001support, roth2006margin} with shrinking threshold.\looseness=-1

\begin{result}[Myopic policy $\ploc_{\mathcal{R},t}$ for the cumulative cost] \label{res:loc_regret}    The myopic policy $\ploc_{\mathcal{R},t}$ is given by
\begin{align}\label{eq:p_loc_opt_regret}
    \ploc_{\mathcal{R},t}(\cdot)=\mathds{1}[|\cdot| \le \kappa^\circ_t],
\end{align}
where the margin $\kappa^\circ_t$ reads
\begin{align}    \label{eq:klocalopt}\kappa^\circ_t=\sqrt{q^\circ_t}\sqrt{2(\sfrac{q^\circ_t}{(m^\circ_t)^2}-1)}\left|\mathrm{erf}^{-1}\left(2\Delta-1\right)\right|,
\end{align}
with $q^\circ_t,m^\circ_t$ satisfying self-consistently the equations \eqref{eq:ODEs} with policy $p_t=\ploc_{\mathcal{R},t}$. As $t\to \infty$, for $\eta=2$, the threshold $\kappa^\circ_t$ shrinks asymptotically as
\begin{align}\label{eq:scaling_kappat}
  \kappa^\circ_t =  \frac{\sqrt{q_0}\,c(\Delta)}{t}+o\left(\frac{1}{t}\right)\,, 
\end{align}
where $c(\Delta)>0$ is given in Eq.~\eqref{eq:c_app} in App.\,\ref{app:Localopt_R}.
\end{result}

\begin{figure*}[t!]
    \centering
    \includegraphics[width=0.4\linewidth]{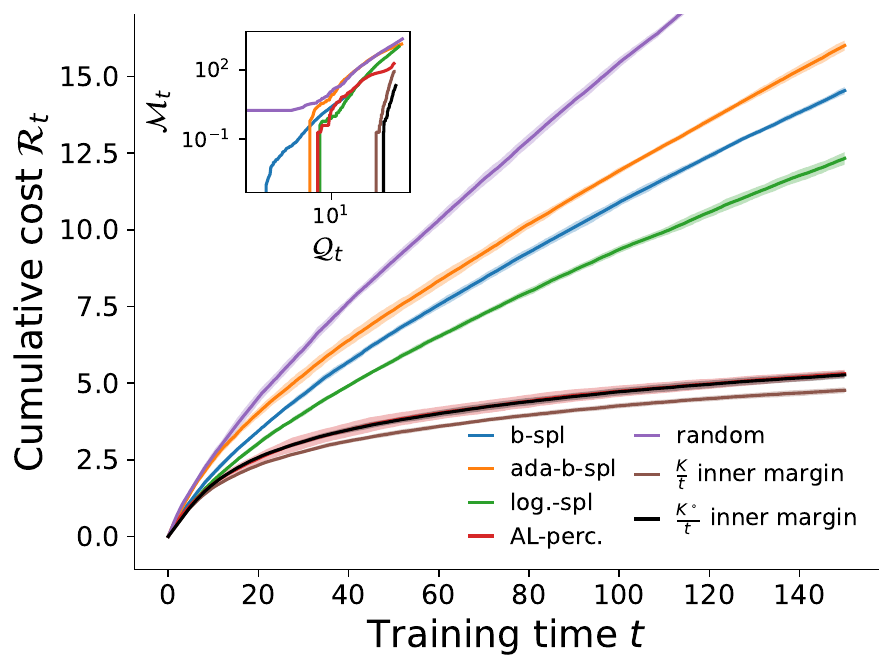}
    \hspace{.8cm}
    \includegraphics[width=0.4\linewidth]{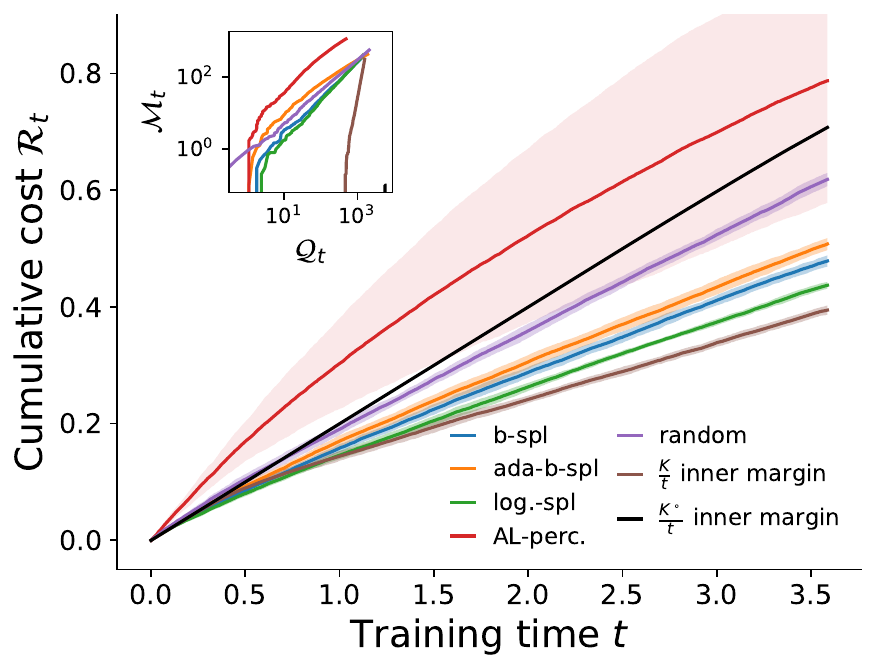}

    \caption{Cumulative cost $\mathcal{R}_t$ and number of queries $\mathcal{Q}_t$ versus mistakes $\mathcal{M}_t$ (inset) for the first-order and adaptive b-sampling algorithms \citep{cesa2004worst} (blue, orange), the logistic sampling algorithm \citep{sculley2007online} (green) and the active perceptron algorithm \citep{dasgupta2005analysis} (red), for binary classification on isotropic Gaussian data in dimension $d=200$ (\textbf{left}), and the pneumonia diagnosis task on chest X-ray images \citep{kermany2018identifying} (\textbf{right}). Purple curves mark a random (constant, margin-independent) strategy. Brown curves represent an inner margin policy with shrinking $\sfrac{K}{t}$ threshold, with tunable bandwidth $K$. The query cost was set to $\Delta=0.2$. Hyperparameters were tuned in hindsight to minimize $\mathcal{R}_{t_F}$, among a grid of $50$ values. The black curve represents the simple thresholding strategy $\ploc_t(\cdot)=\mathds{1}[|\cdot|\le \sfrac{K^\circ}{t}]$, where $K^\circ\approx 2.52$ is the asymptotically optimal value prescribed by \eqref{eq:scaling_kappat}  for $q_0=0.1, \eta=2, \Delta=0.2$. Shades mark one standard deviation over $5$ trials, with the dataset reshuffled at each run. }
    \label{fig:Benchmark_Gaussian}
\end{figure*}

As in Sec.~\ref{sec:budget}, we find a deterministic policy because the instantaneous objective is linear in $p_t$. This simple policy notably induces a sub-linear growth for cost metrics, and logarithmic label complexity, as formalized in the following remark.

\begin{remark}\label{rem:scalings_locopt_QM}
  Under the myopic policy $\ploc_{\mathcal{R},t}$  characterized in Result \ref{res:loc_regret}, for $\eta=2$, the cumulative cost $\mathcal{R}^\circ_t$, number of mistakes $\mathcal{M}^\circ_t$ and number of queries $\mathcal{Q}^\circ_t$ scale logarithmically with training time: 
  \begin{align}
      \mathcal{R}_t^\circ, \mathcal{M}_t^\circ, \mathcal{Q}_t^\circ \asymp \ln t\,.
  \end{align}
  The error scales as $\varepsilon_{g,t}^\circ \asymp \sfrac{1}{t}$, and thus decays exponentially with the number of queries $\mathcal{Q}_t^\circ$.  These scalings match for example those reported in \cite{yan2017revisiting,huang2022disagreement, cavallanti2008linear}.
\end{remark}

\subsection{Horizon-optimized intermediate-margin policy \texorpdfstring{$\pglob_{\mathcal{R},t}$}{}}

We now turn to the horizon-optimized policy $\pglob_{\mathcal{R},t}$ for the cumulative cost, again focusing on the class of intermediate-margin strategies, as for Res.\,\ref{res:pglob_eg}.

\begin{result}[Horizon-optimized policy for the cumulative cost]
We again optimize an intermediate-margin policy
\begin{equation}
\pglob_{\mathcal{R},t}(\cdot)
        =\mathds{1}[a^\star_t\le|\cdot| \le b^\star_t]\;,
\end{equation}    
  where the optimal thresholding schedule is given by
\begin{equation}\label{eq:globally_R}
(a_t^\star,b_t^\star) = \operatorname{argmin}_{\{a_t,b_t\}} \mathcal{R}_{t_F}
        \,.
\end{equation}
For all parameter values considered, we obtain $a^\star_t=0$, so that the horizon-optimized strategy collapses onto an inner-margin policy. This collapse is shown in Fig.~\ref{fig:sel_R} (right) and largely extends to time independent policies.
\end{result}
Fig.~\ref{fig:sel_R} (right) shows that time-dependent schedules achieve lower cost than the best random policy $p_t(\cdot)=p$ and all constant strategies. All free parameters, including $p$, are chosen to minimize $\mathcal{R}_{t_F}$. The horizon-optimized schedule outperforms the myopic one, with a gap growing with $\Delta$. The myopic rule
queries only when it lowers the cost immediately, ignoring that each query also reduces future mistakes. At large $\Delta$ it therefore queries too little, whereas the horizon-optimized schedule invests in queries early in training that pay off later on.

\subsection{Experiments}
The particularly compact asymptotic form \eqref{eq:scaling_kappat} for the myopic threshold \eqref{eq:p_loc_opt_regret} naturally inspires the simple heuristic policy schedule $p_t(\cdot)=\mathds{1}[|\cdot|\le \sfrac{K}{t}],$ with a threshold shrinking at a $\sfrac{1}{t}$ rate. We compare the performance of this policy against other margin-based rules for online Adatron/Perceptron learning proposed in the literature. \cite{cesa2004worst} --blue lines in Fig.\,\ref{fig:Benchmark_Gaussian}-- (resp. \cite{sculley2007online} --green lines in Fig.\,\ref{fig:Benchmark_Gaussian}--) propose a sigmoid (resp. exponential)  functional ansatz for the selection probability $p_t(\cdot)$. Orange lines represent a variant of the former rule, with adaptive bandwidth \citep{cesa2004worst}. \cite{dasgupta2005analysis} --red lines in Fig.\,\ref{fig:Benchmark_Gaussian}-- on the other hand rely on a hard inner-margin selection rule $p_t(\cdot)=\mathds{1}[|\cdot|\le \kappa_t]$ with an adaptive threshold that halves after every streak of consecutive correct predictions. Details on these algorithms and their implementation are collected within App.~\ref{app:numerics}. 
Fig.\,\ref{fig:Benchmark_Gaussian} (left) illustrates the time evolution of the cumulative cost $\mathcal{R}_t$ for these different policies, presenting only the choice of hyperparameter that minimizes in hindsight the terminal cost $\mathcal{R}_{t_F}$. The heuristic  $p_t(\cdot)=\mathds{1}[|\cdot|\le \sfrac{K}{t}]$ inspired from the $t\to\infty$ asymptotics of Result \ref{res:loc_regret} --brown lines in Fig.\,\ref{fig:Benchmark_Gaussian}-- compares favorably against the aforementioned baselines, and results in overall small cumulative cost $\mathcal{R}_t$. The performance for the particular prescription $K^\circ=\sqrt{q_0}\,c(\Delta)$ for the hyperparameter $K$, which can be read from the theoretical expression \eqref{eq:scaling_kappat}, is on the other hand found to deteriorate at larger costs $\Delta$, as illustrated in Figs.\,\ref{fig:Benchmark_Chest} and \ref{fig:otherDelta_Gauss} in App.\,\ref{app:numerics}.
\begin{wrapfigure}{r}{0.38 \linewidth}
    \centering
    \includegraphics[width=0.99\linewidth]{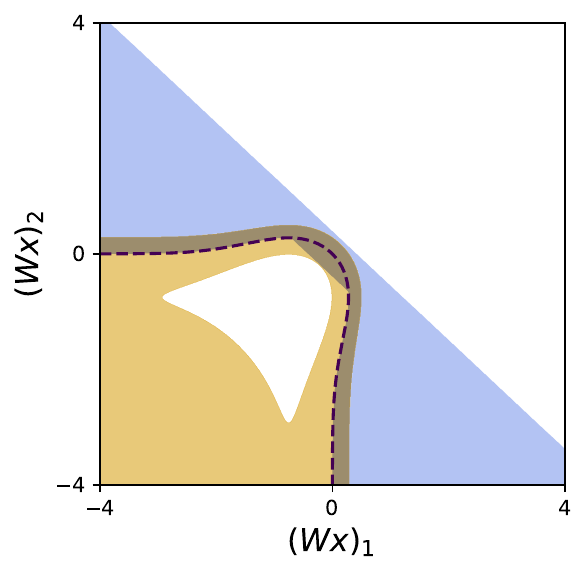}
\caption{Preactivations $Wx$ selected by the myopic policy $\ploc_{\mathcal{R},t}$ \eqref{eq:opt_p_2LNN} (blue) for $\Delta=0.46$ and a naive inner-margin threshold on the network output $p_t(\cdot)=\mathds{1}[|v^\top g(\cdot)|\le 0.35]$ (yellow). The weights $W$ in  \eqref{eq:opt_p_2LNN} were extracted from a GeLU-activated two-layer neural network with $r=2$ hidden neurons, trained without data selection on the logistic loss for $1000$ steps, at $d=500$ and $\eta=1$. The untrained readout is $v=(1,1)^\top$. The decision boundary of the network is represented in dashed lines.  }
    \label{fig:2LNN}\vspace{-7mm}
\end{wrapfigure}
This observation stands in coherence with the discussion of the previous subsection that the performance of the myopic policy comparatively degrades at larger $\Delta$, see Fig.\,\ref{fig:sel_R} (right). A finer examination of the joint evolution of $\mathcal{M}_t, \mathcal{Q}_t$ (Fig.\,\ref{fig:Benchmark_Gaussian}, inset) reveals that the shrinking threshold $\sfrac{K }{t}$ incentivizes a first phase of exploration, with a sharp increase of the number of queries $\mathcal{Q}_t$, followed by a phase of exploitation where the model attempts more predictions, as signaled by a sharp increase in mistakes $\mathcal{M}_t$. This qualitative behavior can still be observed for real data that does not satisfy the Gaussian assumption, as shown in Fig.\,\ref{fig:Benchmark_Gaussian} (right) for the task of pneumonia diagnosis from chest X-ray images \citep{kermany2018identifying}, and in Fig.~\ref{fig:otherDelta_Covid} in App.~\ref{app:numerics} for Covid-19 diagnosis \citep{chowdhury2020can, rahman2021exploring}. \looseness=-1

\subsection{Comments for two-layer neural networks}
The findings of Sec.~\ref{sec:integrated_error} make it tempting to wonder whether output-thresholding strategies are generically well-adapted to control the cumulative cost $\mathcal{R}_t$ \eqref{eq:Rk}. This final section nuances this picture, and shows that this intuition ceases to hold for more complex learning models. Consider for instance the width-$r$ two-layer neural network $
    f(\cdot)= v^\top g\left(W \cdot\right)$
with weights $v\in \R^r, W\in \R^{r\times d}$, and activation function $g$ acting entry-wise. We again consider training this model on the classification task described in Sec.~\ref{sec:setup}, using online SGD \eqref{eq:discrete_GD_step}. 
Note that when $r=\Theta_d(1)$ as $n\asymp d$ diverge, the resulting training dynamics are again captured by a limiting set of low-dimensional ODEs \citep{saad1995exact, goldt2019dynamics}.
For such models, it is natural to consider the larger class of selection policies $p_t: \R^r \to [0,1]$ that take as argument for a given input $x$ the \textit{whole array of pre-activations} $Wx$, rather than the sole output $f(x)$. We note that policies bearing exclusively over the output in fact correspond to special instances of the former class. Is the myopic policy for the cumulative cost still of the form of a hard-thresholding \eqref{eq:p_loc_opt_regret} on the output $f(x)$? The next Result, derived in App.\,\ref{app:Localopt_R}, answers this question negatively.\looseness=-1

\begin{result}[Myopic policy for two-layer networks] \label{res:2LNN_popt}Conditional on the weights $v, W$ at time $t$, supposing $W$ is full rank and $\beta \notin {\rm rowsp}(W)$, the myopic policy for the cumulative cost at time $t$ is given by
    \begin{align}\label{eq:opt_p_2LNN}
    \hspace{-.9em}
    \ploc_{\mathcal{R},t}(\cdot)=\mathds{1}\left[
    \Phi\left(
    \frac{-\mathrm{sign}(v^\top g(\cdot)) M^\top Q^{-1} \cdot }{\sqrt{1-M^\top Q^{-1}M}}
    \right)\ge \Delta
    \right],
\end{align}
where $M\coloneqq\sfrac{W\beta}{d}\in\R^r$ and $Q\coloneqq\sfrac{WW^\top}{d}\in \R^{r\times r}$. $\Phi(\cdot)$ is the normal cumulative distribution function.
\end{result}
While the optimal policy is still found to be deterministic, it no longer depends on the pre-activation $Wx$ through the sole output $v^\top g(Wx)$. In fact, the regions in pre-activation space that the (oracle) policy \eqref{eq:opt_p_2LNN} prescribes to select can significantly differ from those selected by a naive output-thresholding strategy of the form $\mathds{1}[|v^\top g(\cdot)|\le \kappa]$. Fig.\,\ref{fig:2LNN} shows, for a $g=\mathrm{GeLU}$ network with $r=2$ hidden neurons trained without data selection on the logistic loss for $1000$ steps,  that the two regions (respectively in blue and yellow) might in fact barely overlap.

\section*{Conclusions}
We investigate data selection schedules for online active learning with linear classifiers. In the high-
dimensional realizable Gaussian setup, we precisely
characterize the margin-based selection policy schedules that minimize at each round the increment in error or cost of mistakes and queries, down to the constant. In many instances, we find that
optimal strategies coincide with deterministic inner-
or intermediate-margin selection rules with shrinking
thresholds. Within these classes of policies, we further numerically determine the optimal threshold schedules that minimize the final error under a labeling budget
or the cumulative cost. Finally, we exemplify how optimal selection policies for two-layer neural networks can take a more complex form. We believe that a precise investigation of the latter constitutes an interesting avenue for future research.\looseness=-1

\subsection*{Limitations and perspectives}
The theoretical contributions of the present work primarily describe linear classifiers, and moreover rest on strong assumptions, notably that of an isotropic Gaussian data distribution. It would be interesting to extend our analysis to models in which the input distribution itself encodes task-relevant structure, such as the hidden manifold model \citep{goldt2020modeling} or class-dependent Gaussian mixtures \citep{pmlr-v139-refinetti21b}, in order to investigate how this additional information affects query selection. Furthermore, we restricted ourselves to query policies that are functions of the sample margin (equivalently, model output) $\inprod{w_i, x_i}$. While this class encompasses many heuristics in the literature \citep{cesa2004worst, sculley2007online}, there exist functionally more intricate policies that do not fall therein, such as the second-order rule of \cite{cesa2004worst} (Fig.\,3). Finally, while we focused on a single stationary task, a promising direction would be to investigate how to allocate a limited labeling budget across tasks in multi-task settings. Building on \cite{mori2025optimal}, this could clarify how task similarity and interference affect optimal labeling strategies at different stages of learning.

\subsection*{Acknowledgements}

H.~Cui is grateful to Richard Combes and Mo\"ise Blanchard for introducing him to the selective sampling problem. F.~Mignacco is grateful to the Center for Brain Science at Harvard University for its hospitality during her visit, when this work was conducted. F.~Mori was supported by the Center of Mathematical Sciences and Applications
(CMSA) at Harvard University. F.~Mignacco was supported by the Human Frontier Science Program (award number: LT0049/2023-L).

\subsection*{AI use statement}

In this work, we used generative AI tools to assist with the implementation of methods, specifically in developing the numerical code used to produce Figs.~\ref{fig:sel_epsilon}--\ref{fig:sel_R}, \ref{fig:concentration} and \ref{fig:grid}. We did not use generative AI tools to develop theoretical models or conceptual frameworks, formulate mathematical claims, provide critical ingredients for proving mathematical claims, assist in the writing of proofs, propose or refine hypotheses, design or provide feedback on research methodology or experiments, clean or reformat datasets, support qualitative or thematic data analysis, or interpret results. Generating synthetic datasets and assisting with translation were not applicable to this work.

Additionally, we used generative AI tools to draft an initial version of App.~\ref{app:control} and for minor rewriting and stylistic improvements to the manuscript. We reviewed all AI-assisted work and, in particular, verified and tested the AI-assisted code for correctness. We take responsibility for the final content of this work, including all text, claims, and artifacts produced with the aid of generative AI.

\bibliography{references.bib}
\bibliographystyle{unsrtnat}


\clearpage
\appendix
\thispagestyle{empty}

\onecolumn

\section{Derivation of the ODEs}\label{app:ODEs}

In this appendix, we derive the ordinary differential equations (ODEs) for the order parameters $m$ and $q$ given in Eq.~\eqref{eq:ODEs} of the main text, tracking the learning dynamics of online SGD:
\begin{align}
    \label{eq:app_online_update}
        w_{i+1}=w_i-\eta \sigma_i\ell^\prime(\inprod{w_i,x_i}y_i)y_i x_i,
    \end{align}
    with learning rate $\eta>0$. The teacher $\beta\in\mathbb{R}^d$ is fixed throughout the training, with $\Vert \beta\Vert^2_2=d$, and the labels are $y_i=\sgn(\inprod{\beta,x_i})$. At each step $i$, the selection process is encoded by the Bernoulli variable $\sigma_i\sim\mathcal{B}(p_i(\inprod{w_i,x_i}))$. In the following, we will consider differentiable losses $\ell:\R\to \R_+$ of the signed margin, highlighting the Adatron rule with $\ell(z)=\sfrac12\min(z,0)^2$ as a special case. 
    
    In the online setup, at each step $i$ a fresh sample $x_i\sim\mathcal{N}(0_d,I_d/d)$ is considered, independent of the weights $w_i$. Therefore, conditional on $w_i$, the margins $(\omega^*_i,\omega_i)=(\inprod{\beta,x_i},\inprod{w_i,x_i})$ are also jointly Gaussian with zero mean and second moments given by:
    \begin{align}
 \label{eq:app_op_def}\mathbb{E}\left[(\omega_i^*)^2\right]=\frac{\inprod{\beta,\beta}}{d}=1\;,
            &&\mathbb{E}\left[\omega_i^2\right]=\frac{\inprod{w_i,w_i}}{d}\coloneqq q_i\;,
            &&\mathbb{E}\left[\omega_i^*\,\omega_i\right]=\frac{\inprod{\beta,w_i}}{d}\coloneqq m_i\,,
    \end{align}
 where the expectations are over $x_i$, conditional on $w_i$. We have defined the order parameters $m_i$ (alignment with the teacher) and $q_i$ (squared norm of the weights), which fully determine the joint conditional  margin distribution
 \begin{equation}
(\omega^*,\omega)\sim\mathcal{N}\left(\begin{bmatrix} 0\\ 0\end{bmatrix}, \begin{bmatrix} 1 & m\\ m & q\end{bmatrix}\right) \,.
 \end{equation}
 The generalization error depends on the weights only through the margins $(\omega^*,\omega)$
 and can thus be expressed solely in terms of the correlation coefficient $\sfrac{m}{\sqrt{q}}$ 
\begin{equation} \label{eq:eg}
    \varepsilon_g = \P{\omega \,\omega^*<0}=\frac{1}{\pi}\arccos\left(\frac{m}{\sqrt{q}}\right)\,.
\end{equation}
Therefore, in order to track the generalization error over training, it is sufficient to track the dynamics of the order parameters. Substituting the update rules \eqref{eq:app_online_update} into the definitions \eqref{eq:app_op_def} gives the exact increments 
\begin{equation}
\label{eq:app_op_update}
        m_{i+1}-m_i=-\frac{\eta} d\sigma_i\ell'(\omega_i y_i)y_i\omega^*_i\;,\qquad
        q_{i+1}-q_i=-2\frac\eta d\sigma_i\ell'(\omega_i y_i)y_i\omega_i+\frac{\eta^2}d\sigma_i(\ell'(\omega_i y_i))^2\inprod{x_i,x_i}\;,
\end{equation}
 where we have used that $\sigma_i\in\{0,1\}$, hence $\sigma_i^2=\sigma_i$. The quadratic term in the increment of $q$ must be retained, since it contributes at the same order in $\sfrac 1d$ as the linear term. 
 
 We take the proportional infinite-dimensional limit $d,n\rightarrow\infty$ with $t_F=\sfrac{n}{d}$ fixed and consider the macroscopic time $t=\sfrac i d\in[0,t_F]$, where each increment is of order ${\rm d}t=\sfrac{1}{d}$. The learning rate $\eta$ is also held fixed. 
The input norm satisfies $
\mathbb E[\inprod{x_i,x_i}]=1$ and
$\operatorname{Var}(\inprod{x_i,x_i})=\sfrac{2}{d}$, and hence it converges in probability: $\inprod{x_i,x_i}\xrightarrow[d\to\infty]{\mathbb P}1$. For an isotropic Gaussian initialization $w_0\sim\mathcal N(0,q_{\mathrm{init}}I_d)$ independent of the teacher, the initial order parameters also concentrate as $d\to\infty$. Indeed, averaging over the initialization and using $\|\beta\|^2=d$,
\begin{equation}
\mathbb E[m_0]=0,\qquad
\operatorname{Var}(m_0)=\frac{q_{\mathrm{init}}}{d},
\qquad
\mathbb E[q_0]=q_{\mathrm{init}},\qquad
\operatorname{Var}(q_0)=\frac{2q_{\mathrm{init}}^2}{d}.
\end{equation}
Consequently, $(m_0,q_0)\xrightarrow[d\to\infty]{\mathbb P}(0,q_{\mathrm{init}})$, providing deterministic initial conditions for the limiting dynamics.

 To control the increments during training, we consider losses that satisfy $|\ell'(z)|\le C(1+|z|)$ for all $z\in\mathbb R$, with a finite constant $C>0$ independent of $z$ and $d$. This encompasses squared hinge, square, and logistic losses. The increments in Eq.~\eqref{eq:app_op_update} scale as $d^{-1}$.  Over $i=d\,t$ steps at fixed $t$, their conditional means generate a finite drift, while the centered fluctuations are uncorrelated across steps and have cumulative standard deviation of order $d^{-1/2}$, vanishing as $d\to\infty$. Here, conditioning is on the training history before step $i$, and the averages are over the fresh input $x_i$ and query decision $\sigma_i$.

 The limiting dynamics are therefore determined by the conditional mean increments, following the standard ODE approach to online
learning \cite{saad1995exact,biehl1995learning,engel2001statistical} (see, e.g., \citep{goldt2019dynamics,arous2021online} for a more rigorous treatment). Averaging over $\sigma_i$ replaces it by $p_i(\omega_i)$, while
input-norm concentration allows to replace $\inprod{x_i,x_i}$ by one. Dividing the resulting mean increments by $\dd t=\sfrac 1d$, the order parameters dynamics \eqref{eq:app_op_update} concentrate in probability to their deterministic limit 
 \begin{equation}
 \begin{split}
     \frac{\dd m}{\dd t}&=f_m(m_t,q_t;p_t)=-\eta\mathbb{E}[p_t(\omega)\ell'(\omega y)y\omega^*]\;,\\
      \frac{\dd q}{\dd t}&=f_q(m_t,q_t;p_t)=-2\eta\mathbb{E}[p_t(\omega)\ell'(\omega y)y\omega]+\eta^2 \mathbb{E}[p_t(\omega)(\ell'(\omega y))^2].
      \label{eq:app_ODEs}
\end{split}
 \end{equation}

 \begin{figure}
\centering
\includegraphics[width=0.45\linewidth]{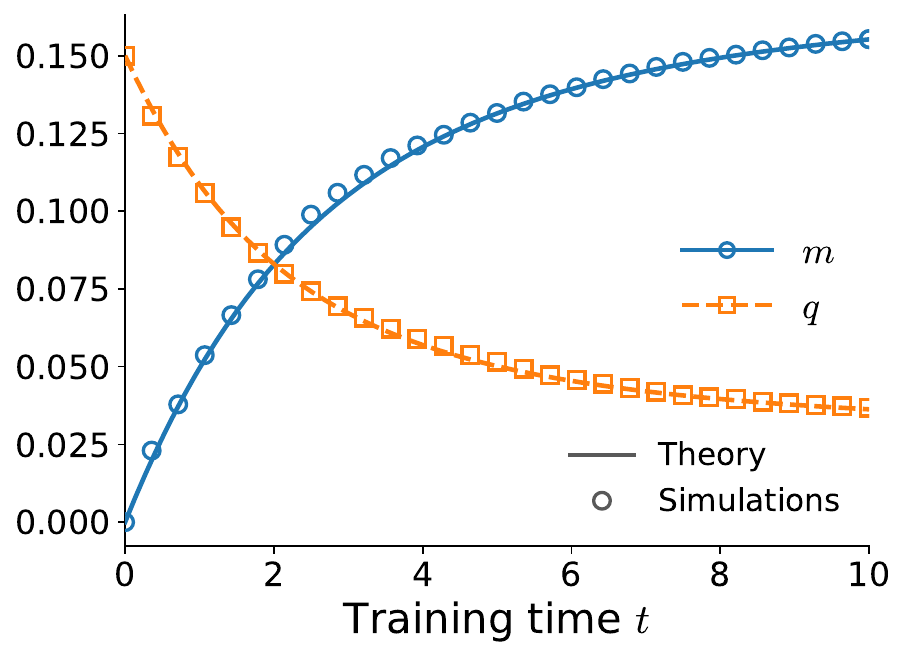}
  \caption{Summary statistics $m$ and $q$ as a function of training time $t$, for the constant policy $p_t(\omega)=\mathds{1}_{|\omega|<1}$.  Lines are obtained by numerically integrating the ODEs \eqref{eq:ODEs}, while symbols mark numerical simulations of a single SGD trajectory at $d=1000$. Parameters:
    $\eta=0.5$, $m_0=0$, $q_0=0.15$, $t_F=10$.}
  \label{fig:concentration}
\end{figure}
At a fixed training time $t$, we write $m=m_t$ and $q=q_t$. Since the policy depends only on the margin $\omega$, it is convenient to first average over the conditional teacher margin
$\omega^*\mid\omega\sim\mathcal N(\mu,s^2)$, where $\mu=(\sfrac m q) \omega$ and $s=\sqrt{1-m^2/q}$. Denoting the standard normal density (resp.~cumulative distribution function) as $G$ (resp.~$\Phi$), the marginal density of $\omega$ as $P_q$, and its standardized conditional mean as $\zeta$, 
\begin{align}
      G(x)=\frac{e^{-x^2/2}}{\sqrt{2\pi}}, \quad
  \Phi(x)=\int_{-\infty}^{x}\!G(u)\,\dd u, \quad
  P_q(\omega)=\frac{e^{-\omega^2/(2q)}}{\sqrt{2\pi q}}, \quad
  \zeta=\frac{\mu}{s}=\frac{m\,\omega}{\sqrt{q(q-m^2)}}\,,
  \label{eq:app_zeta}
\end{align}
the three moments appearing in Eq.~\eqref{eq:app_ODEs} are given by
\begin{align}
  \E[(\ell'(\omega y))^2\mid\omega] &= \Phi(\zeta)\,\ell'(\omega)^2 + \Phi(-\zeta)\,\ell'(-\omega)^2, \\
  \E[\ell'(\omega y)y\,\omega\mid\omega] &= \omega\big[\Phi(\zeta)\,\ell'(\omega)
     - \Phi(-\zeta)\,\ell'(-\omega)\big], \\
  \E[\ell'(\omega y)y\,\omega^\ast\mid\omega] &= \ell'(\omega)\big(\mu\Phi(\zeta)+sG(\zeta)\big)
     - \ell'(-\omega)\big(\mu\Phi(-\zeta)-sG(\zeta)\big).
  \label{eq:app_moments}
\end{align}
For any function $h(\omega,\omega^*)$, the corresponding term in \eqref{eq:app_ODEs} factorizes as 
\begin{equation}
    \mathbb{E}[p_t(\omega) h(\omega,\omega^*)]=\int_{-\infty}^\infty \dd \omega \, P_q(\omega)\, p_t(\omega)\,\mathbb{E}[h(\omega,\omega^*)|\omega]\;.
\end{equation}
All three moments in \eqref{eq:app_moments} are even in $\omega$, so every policy that we derive in the following can be written as a condition on $|\omega|$ alone.

For the Adatron rule, $\ell'(z)=\min(z,0)$, so $y\ell'(\omega y)=\omega\mathbf{1}_{\{\omega\omega^*<0\}}$ and $(\ell'(\omega y))^2=\omega^2\mathbf{1}_{\{\omega\omega^*<0\}}$. Each nonzero update increases the overlap $m$, therefore,
starting from $m_0=0$, we have $m\ge0$ nondecreasing throughout
training. Using the conditional
misclassification probability
$\mathbb{P}[\omega\omega^*<0\mid\omega]=\Phi(-|\zeta|)$
for $\omega\ne0$, Eq.~\eqref{eq:app_ODEs} becomes
\begin{align}
\label{eq:app_adatron_ODEs}
    \begin{split}
f_m(m_t,q_t;p_t)
&=\eta\int_{-\infty}^\infty\dd\omega\,P_{q_t}(\omega)p_t(\omega)
\left[
s_t|\omega|G(\zeta_t)
-\frac{m_t}{q_t}\omega^2\Phi(-|\zeta_t|)
\right]\,,\\
f_q(m_t,q_t;p_t)
&=\eta(\eta-2)\int_{-\infty}^\infty\dd\omega\,
P_{q_t}(\omega)p_t(\omega)\omega^2\Phi(-|\zeta_t|)\,.
    \end{split}
\end{align}
The self-overlap $q_t$ is nonincreasing for $0<\eta<2$ and remains constant for $\eta=2$
in the limiting dynamics.

Figure \ref{fig:concentration} compares the ODE predictions \eqref{eq:app_adatron_ODEs} with numerical simulations of a single SGD trajectory at $d=1000$ under the fixed policy $p_t(\omega)=\mathbf{1}_{\{|\omega|<1\}}$. During training, the overlap $m_t$ increases while the normalized squared weight norm $q_t$ decreases, increasing their correlation $m_t/\sqrt{q_t}$ and thereby reducing the generalization error. The close agreement between the theoretical curves and simulation points shows that the ODEs accurately capture this evolution already at finite dimension.

Similarly, we derive the dynamics of the (rescaled) cumulative cost ${\cal R}_k={(\mathcal{M}_k+\Delta \,{\cal Q}_k)}/d$ (see Eq.~\eqref{eq:Rk} of the main text) by averaging its one-step increments
over the fresh input $x_i$ and query decision $\sigma_i$,
conditional on the training history before step $i$. We recall that ${\cal Q}_k=\sum_{i=1}^k\sigma_i$ denotes the cumulative number of queries, and ${\cal M}_k=\sum_{i=1}^k(1-\sigma_i)\mathds{1}[\omega_iy_i<0]$ the cumulative number of mistakes. Since $d$ samples arrive per unit of macroscopic time, with a slight abuse of notation here we denote
by ${\cal Q}_t$ and ${\cal M}_t$ the deterministic limits of
$\mathcal{Q}_k/d$, $\mathcal{M}_k/d$ at $k=\lfloor d\,t\rfloor$, while $\mathcal{R}_t=\mathcal{R}_{\lfloor d\,t\rfloor}$ as in the main text. Their time evolution is given by 
\begin{equation}
\label{eq:app_cost_dynamics}
\begin{split}
{\cal Q}_t
&=\int_0^t \dd t'\int_{-\infty}^\infty\dd\omega\,P_{q_{t'}}(\omega)p_{t'}(\omega)\,,\\
{\cal M}_t
&=\int_0^t \dd t'\int_{-\infty}^\infty\dd\omega\,P_{q_{t'}}(\omega)
[1-p_{t'}(\omega)]\Phi(-|\zeta_{t'}|),\\
{\cal R}_t
&={\cal M}_t + \Delta\,{\cal Q}_t\,,
\end{split}
\end{equation}
The constraint that a fraction $\gamma$ of the samples
is queried becomes 
\begin{equation}
\frac{{\cal Q}_{t_F}}{t_F}
=\frac{1}{t_F}\int_0^{t_F}\dd t
\int_{-\infty}^\infty\dd\omega\,P_{q_t}(\omega)p_t(\omega)
=\gamma\,.
\end{equation}

\section{Control methods}\label{app:control}

\begin{figure}
    \centering
    \includegraphics[width=\linewidth]{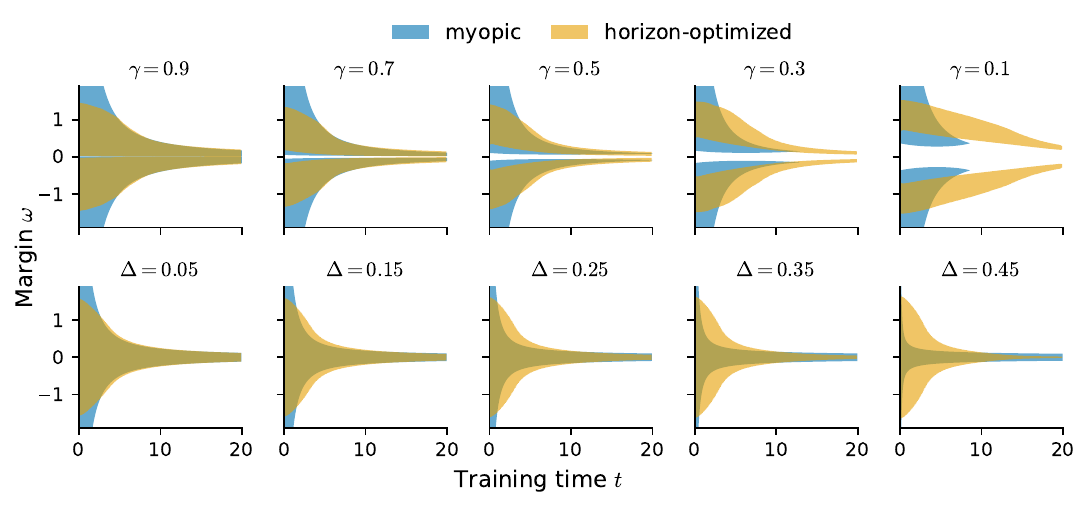}
      \caption{Selection policies in the $(t,\omega)$ plane for final-error (top row) and cumulative-cost (bottom row) minimization. The colored regions indicate
  $p_t(\omega)=1$. Parameters: $\eta=0.5$, $m_0=0$, $q_0=0.15$, $t_F=20$.}
    \label{fig:grid}
\end{figure}

The ODEs of Result~\ref{res:ODEs} reduce the search for a selection policy to a
low-dimensional optimal control problem. The state is $(m_t,q_t)$, the control
is the policy $p_t$, and the objective is either the final error $\varepsilon_{g,t_F}$
under the budget \eqref{eq:budget}, or the cumulative cost $\mathcal{R}_{t_F}$.
This appendix describes how we solve this problem. We first derive the myopic
policy, then explain how we discretize and optimize time-dependent schedules.

\subsection{Myopic policies}

Both $f_m$ and $f_q$ are linear in $p_t$, so the rate of change of the error is
linear in $p_t$ as well. We can write it as
\begin{equation}
  \frac{\dd\eg}{\dd t}
  = \E\big[p_t(\omega)\,S_t(\omega)\big],
  \qquad
  S_t(\omega)
  = \frac{\partial\eg}{\partial m}\,
      \frac{\delta f_m}{\delta p_t(\omega)}
  + \frac{\partial\eg}{\partial q}\,
      \frac{\delta f_q}{\delta p_t(\omega)},
  \label{eq:slope}
\end{equation}
where $\partial\eg/\partial m=-1/(\pi\sqrt{q-m^2})$ and
$\partial\eg/\partial q=m/(2\pi q\sqrt{q-m^2})$ follow from \eqref{eq:eg}. We recall that $\E$ denotes the expected value with respect to $\omega\sim \mathcal{N}(0,q)$.
The function $S_t(\omega)$ measures how much querying a sample at margin
$\omega$ changes $\dd\eg/\dd t$. Samples with $S_t(\omega)<0$ reduce the error,
and the others do not. A greedy policy that minimizes \eqref{eq:slope} at each
time would therefore query every $\omega$ with $S_t(\omega)<0$. For the Adatron
rule, however, $S_t$ is negative for all $\omega>0$. The greedy policy then
queries every sample and violates the budget \eqref{eq:budget}.

To enforce the budget, we introduce a Lagrange multiplier $\nu\ge 0$. This is a
single scalar for the whole trajectory, and it acts as a fixed price per query. The instantaneous objective $\mathbb{E}[p_t(\omega)(S_t(\omega)+\nu)]$ is linear in $p_t$. Therefore, optimizing over arbitrary selection probabilities yields a {\it deterministic} myopic schedule almost surely: a sample at margin
$\omega$ is queried only if $S_t(\omega)+\nu<0$, that is, only if the error
reduction it brings outweighs the price of the query, giving the policy of Result~\ref{res:loc_budget}. We tune $\nu$ by bisection
so that the realized query rate equals $\gamma$.

\subsection{Quadrature on the selected intervals}

All the policies we consider select a union of intervals in $|\omega|$, and the
integrands in $f_m$ and $f_q$ are even in $\omega$. For an intermediate-margin
policy $p_t(\omega)=\mathds{1}[a_t\le|\omega|\le b_t]$, every expectation is
therefore an ordinary integral over the selected interval:
\begin{equation}
  \E\big[p_t(\omega)\,X(\omega)\big]
  = 2\int_{a_t}^{b_t}\!\dd\omega\;
      P_{q_t}(\omega)\,X(\omega),
  \qquad
  P_{q}(\omega)=\frac{e^{-\omega^2/2q}}{\sqrt{2\pi q}}\,,
  \label{eq:interval_integral}
\end{equation}
where $X$ stands for any of the conditional moments of
App.~\ref{app:ODEs}. We compute these integrals with an $n$-point
Gauss--Legendre rule, rescaled from $[-1,1]$ to $[a_t,b_t]$:
\begin{equation}
  \omega_j = \frac{a_t+b_t}{2} + \frac{b_t-a_t}{2}\,\xi_j,
  \qquad
  \E\big[p_t(\omega)\,X(\omega)\big]
  \simeq (b_t-a_t)\sum_{j=1}^{n} W_j\,P_{q_t}(\omega_j)\,X(\omega_j),
  \label{eq:gauss_legendre}
\end{equation}
where $\{\xi_j\}$ and $\{W_j\}$ are the standard Gauss--Legendre nodes and
weights on $[-1,1]$. This choice has two advantages. First, it is very accurate. The integrand $P_{q_t}X$ is analytic on
$[a_t,b_t]$, so the quadrature error decreases geometrically with $n$ and we use
$n=48$. Second, and more importantly, it makes the discretized ODE in Eq.~\eqref{eq:gauss_legendre} differentiable with respect to the policy parameter $a_t$ and $b_t$. 

\subsection{Piecewise-constant schedules}

For simplicity, we approximate the schedules to be piecewise constant in time. We
split the horizon into $N_c$ equal blocks and keep the radii fixed within each
block:
\begin{equation}
  a_t = a_k, \quad b_t = b_k
  \qquad\text{for}\quad
  t\in\Big[\tfrac{(k-1)t_F}{N_c},\,\tfrac{k\,t_F}{N_c}\Big),
  \qquad k=1,\dots,N_c.
\end{equation}
The control is then a vector of $2N_c$ numbers, or $N_c$ numbers for
inner-margin policies, where $a_k\equiv 0$. We integrate the ODEs
\eqref{eq:ODEs} with forward Euler on $N_s$ time steps, with $N_s$ a multiple
of $N_c$. The time integrals in \eqref{eq:budget} and in $\mathcal{R}_{t_F}$
are computed with the trapezoidal rule on the same grid. During optimization we
use $N_c=50$ and $N_s=500$. All reported quantities are then recomputed with
$N_s=2000$.

\subsection{Enforcing the query budget}

During optimization, we impose the budget \eqref{eq:budget} through a quadratic
penalty:
\begin{equation}
  \mathcal{L}
  = \varepsilon_{g,t_F}
  + \rho\left(\frac{1}{t_F}\int_0^{t_F}\!\E[p_t(\omega)]\,\dd t-\gamma\right)^2,
  \qquad \rho=300.
  \label{eq:penalty}
\end{equation}
A penalty does not enforce the constraint exactly, so the optimized schedule
can slightly overspend or underspend the budget. We correct this afterwards by
rescaling all interval widths by a common factor $s$,
$b_k\mapsto a_k+s\,(b_k-a_k)$, with $s$ chosen by bisection so that the query
rate equals $\gamma$. We then recompute the error for the corrected schedule.

\begin{table*}[t]
  \centering
  \small
  \begin{tabular}{@{}>{\raggedright\arraybackslash}p{3.0cm}>{\raggedright\arraybackslash}p{3.8cm}>{\raggedright\arraybackslash}p{3.8cm}>{\raggedright\arraybackslash}p{3.8cm}@{}}
    \toprule
    Strategy
      & Policy $p_t(\omega)$
      & Variables optimized  (Fig.~\ref{fig:sel_epsilon})
      & Variables optimized (Fig.~\ref{fig:sel_R}) \\
    \midrule
    Myopic
      & $\mathds{1}[S_t(\omega)+\nu<0]$
      & one scalar $\nu$, so that the query
        rate equals $\gamma$
      & nothing \\
    \addlinespace
    Inner margin, constant
      & $\mathds{1}[|\omega|\le\kappa]$
      & nothing: the budget fixes $\kappa$
      & one scalar $\kappa$ \\
    \addlinespace
    Inner margin, schedule
      & $\mathds{1}[|\omega|\le\kappa_t]$
      & time-dependent $\kappa_t$ 
      & time-dependent $\kappa_t$ \\
    \addlinespace
    Intermediate margin, constant
      & $\mathds{1}[a\le|\omega|\le b]$
      & one scalar $a$, with $b$ fixed by the budget at each $a$
      & two scalars $a,b$ \\
    \addlinespace
    Intermediate margin, schedule
      & $\mathds{1}[a_t\le|\omega|\le b_t]$
      & time-dependent $a_t$ and $b_t$
      & time-dependent $a_t$ and $b_t$ \\
    \addlinespace
    Random
      & $p$, independent of $\omega$ and $t$
      & nothing: the budget forces $p=\gamma$
      & one scalar $p$ \\
    \bottomrule
  \end{tabular}
  \caption{Policy families compared in Figs.~\ref{fig:sel_epsilon} (right) and
  \ref{fig:sel_R} (right). }
  \label{tab:strategies}
\end{table*}

\subsection{Optimization}

We solve the resulting finite-dimensional problem by direct single shooting:
for a given control vector, we simulate the whole trajectory, evaluate the
objective, and update the controls. The forward simulation is implemented in
\texttt{JAX}, which provides exact gradients with respect to the control vector
by reverse-mode automatic differentiation through the discretized trajectory.
We minimize with L-BFGS-B, using box constraints on the controls. The objective has
several local minima, so we run each optimization from a few initializations,
and we warm-start from the previous solution when sweeping $\gamma$ or
$\Delta$. We report the best local optimum found.

\section{Myopic schedules for the cumulative cost}\label{app:Localopt_R}

We detail in this Appendix the derivations of the myopic policy for the cumulative cost objective $\mathcal{R}_t$ \eqref{eq:Rk}, for a linear classifier (Res.\,\ref{res:loc_regret}) and a two-layer neural network (Res.\, \ref{res:2LNN_popt}).

\subsection{Linear classifier}
We first ascertain the policy schedule $\ploc_{\mathcal{R},t}(\cdot)$ that greedily minimizes the local increment in cost $\sfrac{\dd}{\dd t}\mathcal{R}_t$. At a given step $t$, the  latter reads
\begin{align}
    \frac{\dd\mathcal{R}_t}{\dd t}=\Delta \Ea{p_t(\omega)}+\Ea{(1-p_t(\omega))\mathds{1}[\omega\omega^\ast<0}],
\end{align}
where the expectation bears over 
\begin{align}
    \omega, \omega^\ast \sim \mathcal{N}\left(0_2, \begin{bmatrix}
        q_t &m_t\\
        m_t & 1
    \end{bmatrix}\right).
\end{align}
For a given $\omega\in \R$, taking a functional derivative with respect to $p_t(\omega)$, 
\begin{align}
    \frac{\delta}{\delta p_t(\omega)}\frac{\dd \mathcal{R}_t}{\dd t}&=\frac{e^{-\frac{\omega^2}{2q_t^\circ}}}{\sqrt{2\pi q_t}}\left[\Delta -\Ea{\mathds{1}[\omega\omega^\ast<0]\mid \omega}\right]\\
    & =\frac{e^{-\frac{\omega^2}{2q_t}}}{\sqrt{2\pi q_t}}\left[
    \Delta -\Phi\left(\frac{-\sfrac{m_t}{q_t}|\omega|}{\sqrt{1-\sfrac{m_t^2}{q_t}}}\right).
    \right].
\end{align}
The sign of the derivative is dictated by the term in brackets. If the latter is positive, then the risk increment $\sfrac{\dd \mathcal{R}_t}{\dd t}$ is minimal at $p_t(\omega)=0$. If conversely the bracketed term is negative, then the $\sfrac{\dd \mathcal{R}_t}{\dd t}$ is minimal for $p_t(\omega)=1$. The myopic policy for the cumulative risk is thus 
\begin{align}
\ploc_{\mathcal{R},t}(\omega)=\mathds{1}\left[\Phi\left(\frac{-\sfrac{m_t}{q_t}|\omega|}{\sqrt{1-\sfrac{m_t^2}{q_t}}}\right)\ge \Delta \right]
\end{align}
If $\Delta>\sfrac{1}{2}$, the optimal rule never makes a query, since even a random classifier that always attempts a prediction will achieve, on average, smaller cost increment. If $\Delta\le \sfrac{1}{2}$ on the other hand, $\ploc_{\mathcal{R},t}$ reduces to an adaptive margin-based policy of the form 
\begin{align}
   \ploc_{\mathcal{R},t}(\omega)=\mathds{1}[|\omega| \le     \kappa^\circ_t]
\end{align}
with
\begin{align}\label{eq:expression_kappa}
    \kappa^\circ_t =\sqrt{q_t^\circ}\sqrt{2(\sfrac{q_t}{m_t^2}-1)}\left|\mathrm{erf}^{-1}\left(2\Delta-1\right)\right|.
\end{align}
This recovers equations \ref{eq:klocalopt} of Res.\,\ref{res:loc_regret}. The expression of the threshold $\kappa^\circ_t$ thus depends on the model weights as time $t$, as encapsulated by the summary statistics $m_t, q_t$. When the model is trained with the Adatron rule with $\eta=2$, and guided at each round by the $  \ploc_{\mathcal{R},t}$, an analysis of the ODEs \eqref{eq:ODEs} allows to precisely ascertain the asymptotic behavior of $\kappa^\circ_t$ in the limit of large training times $t\to \infty$. This is the objective of the following subsection.

\subsection{Large-time asymptotics}

We now turn to examining the $t\to \infty$ behavior of the threshold $\kappa^\circ_t$ \eqref{eq:expression_kappa} when the summary statistics $q_t^\circ, m_t^\circ$ are self-consistently governed by the ODEs \eqref{eq:ODEs} under policy schedule $  \ploc_{\mathcal{R},t}$, with $\eta=2$. Let us show in self-consistent fashion that
\begin{align}
    \kappa^\circ_t \asymp \frac{1}{t}.
\end{align}

We accordingly denote $\kappa_t^\circ = \frac{A\sqrt{q_0}}{t}+o(\sfrac{1}{t})$, and aim at determining self-consistently the leading coefficient $A$. Denoting $\gamma_t^\circ=\sfrac{m_t^\circ}{\sqrt{q_t^\circ}}$, we also assume that $
     \sfrac{\kappa_t^\circ }{\sqrt{2q_0(1-\gamma_t^\circ)}} $ admits a non-zero limit as $t\to\infty$, which will be subsequently verified self-consistently. We accordingly introduce $z$ such that
\begin{align}
    \frac{\kappa_t^\circ}{\sqrt{2q_0(1-\gamma_t^\circ)}}\xrightarrow{t\to\infty } z, \label{eq:intro_z}
\end{align}
again to be determined self-consistently.
Under these assumptions, the ODE governing the time evolution of the summary statistics $m_t^\circ$ \eqref{eq:ODEs} simplifies as 
\begin{align}
    \frac{\dd m_t^\circ}{\dd t}&=4\int\limits_0^\infty \frac{e^{-\frac{1}{2}(\omega^\star)^2}}{\sqrt{2\pi}} \int\limits_{-\infty}^\infty \frac{e^{-\frac{1}{2}g^2}}{\sqrt{2\pi}} \mathds{1}[|m_t^\circ\omega^\star+\sqrt{2q_t^\circ(1-\gamma_t^\circ)}g|\le \kappa_t^\circ] \\
    &\qquad\qquad\qquad\qquad\qquad\qquad\mathds{1}\left[-(m_t^\circ\omega^\star+\sqrt{2q_t^\circ(1-\gamma_t^\circ)}g)\ge 0\right]|m_t^\circ\omega^\star+\sqrt{2q_t^\circ(1-\gamma_t^\circ)}g|\omega^\star \dd \omega^\ast \dd g\\
    &=-4\int\limits_0^\infty \frac{e^{-\frac{1}{2}(\omega^\star)^2}}{\sqrt{2\pi}}\omega^\star \int\limits_{\frac{-\kappa_t^\circ-m_t^\circ\omega^\star}{\sqrt{2q_t^\circ(1-\gamma_t^\circ)}}}^{-\frac{m_t^\circ\omega^\star}{\sqrt{2q_t^\circ(1-\gamma_t^\circ)}}} \frac{e^{-\frac{1}{2}g^2}}{\sqrt{2\pi}} (m_t^\circ\omega^\star+\sqrt{2q_t^\circ(1-\gamma_t^\circ)}g) \dd \omega^\ast \dd g\\
    &\approx -4m_t^\circ \int\limits_0^\infty \frac{e^{-\frac{1}{2}(\omega^\star)^2}}{\sqrt{2\pi}}(\omega^\star)^2 \left[\Phi\left(\frac{-m_t^\circ\omega^\star}{\sqrt{2q_t^\circ(1-\gamma_t^\circ)}}\right)-\Phi\left(-z-\frac{m_t^\circ\omega^\star}{\sqrt{2q_t^\circ(1-\gamma_t^\circ)}}\right)\right]\dd \omega^\ast \\
    &\qquad +4\sqrt{2q_t^\circ}\frac{1}{\pi}(1-\gamma_t^\circ)^{\frac{3}{2}}\left[
    1-e^{-\frac{z^2}{2}}+\sqrt{2\pi}z\Phi(-z)
    \right]\\
    &\approx \sqrt{q_0} \phi(z) (1-\gamma_t^\circ)^{\frac{3}{2}},
\end{align}
where we introduced
\begin{align}
    \phi(z)=  \frac{4}{\sqrt{2 \pi}}\left(-\int\limits_0^\infty  x^2\left[\Phi(-x/\sqrt{2})-\Phi(-z-x/\sqrt{2})\right]\dd x +\sqrt{4/\pi}\left[
    1-e^{-\frac{z^2}{2}}+\sqrt{2\pi}z\Phi(-z)
    \right]\right).
\end{align}
On the other hand, for a learning rate $\eta=2$ (an assumption also made e.g. by \cite{dasgupta2005analysis}), the squared norm $q_t^\circ$ remains constant along training. Then, it follows from the identity
\begin{align}
    \frac{\dd(1-\gamma_t^\circ)}{\dd t}=-\frac{1}{\sqrt{q_0}}\frac{\dd m^\circ_t}{\dd t}=-\phi(z) (1-\gamma_t^\circ)^{\frac{3}{2}}
\end{align}
that $1-\gamma_t^\circ$ scales as
\begin{align}\label{eq:scaling_gamma}
    1-\gamma_t^\circ \asymp \frac{4}{\phi(z)^2 t^2}
\end{align}
when $t\to \infty$. As a corollary, the test error thus admits the following asymptotic behavior 
\begin{align}
    \varepsilon_g =\frac{1}{\pi}\arccos(\gamma_t^\circ) = \frac{\sqrt{8}}{\pi |\phi(z)| t}+o\left(\frac{1}{t}\right).
\end{align}
Finally, returning to \eqref{eq:intro_z}, one deduces that $z$ satisfies the self-consistent equation
\begin{align}
    \frac{A|\phi(z)|}{\sqrt{8}}=z,
\end{align}
which provides a first self-consistent equation bearing on $z$.

 Having determined the asymptotics of $m_t^\circ, q_t^\circ$, one is now in a position to ascertain the limiting behavior of $\kappa^\circ_t$, and in doing so close the equations on $A$. From \eqref{eq:expression_kappa} and \eqref{eq:scaling_gamma}, $\kappa^\circ_t$ asymptotically shrinks as
\begin{align}
  \kappa^\circ_t \asymp   \frac{4\sqrt{ q_0}\left|\mathrm{erf}^{-1}\left(2\Delta-1\right)\right|}{|\phi(z)|t},
\end{align}
from which one deduces $A=\sfrac{4\left|\mathrm{erf}^{-1}\left(2\Delta-1\right)\right|}{|\phi(z)|}$.
Returning to the self-consistent equation \eqref{eq:intro_z}, the latter simplifies and gives in closed form 
\begin{align}
    z=\sqrt{2}\left|\mathrm{erf}^{-1}\left(2\Delta-1\right)\right|.
\end{align}
for the myopic rule. Finally, we find
\begin{align}
  \kappa^\circ_t \asymp   \sqrt{ q_0} c(\Delta)\frac1t,
\end{align}
with
\begin{equation}\label{eq:c_app}
c(\Delta)=\frac{4\left|\mathrm{erf}^{-1}\left(2\Delta-1\right)\right|}{|\phi(\sqrt{2}\left|\mathrm{erf}^{-1}\left(2\Delta-1\right)\right|)|}\,.
\end{equation}

\subsection{Derivation of Remark \ref{rem:scalings_locopt_QM}}
The expected regret increment is thus at large time given at leading order by
\begin{align}
    \Ea{\frac{\dd \mathcal{R}_t^\circ}{\dd t}}&=\Delta \Ea{p_t(\omega)}+\Ea{(1-p_t(\omega))\mathds{1}[\omega\omega^\star<0}]\\
    & = \left[2\Phi\left(\frac{\kappa^\circ_t}{\sqrt{q_0}}\right)-1\right]\Delta +2\sqrt{1-\gamma} \frac{1}{\sqrt{2\pi}}\int\limits_0^\infty  \Phi\left(-\sqrt{2}\left|\mathrm{erf}^{-1}\left(2\Delta-1\right)\right|-x/\sqrt{2}\right) \dd x\\
    & \asymp \sqrt{\frac{2}{\pi}}\Delta \frac{4\left|\mathrm{erf}^{-1}\left(2\Delta-1\right)\right|}{|\phi\left(\sqrt{2}\left|\mathrm{erf}^{-1}\left(2\Delta-1\right)\right|\right)|t}+\frac{\frac{4}{\sqrt{2\pi}}\int\limits_0^\infty  \Phi\left(-\sqrt{2}\left|\mathrm{erf}^{-1}\left(2\Delta-1\right)\right|-x/\sqrt{2}\right)\dd x}{ |\phi\left(\sqrt{2}\left|\mathrm{erf}^{-1}\left(2\Delta-1\right)\right|\right)|t}.
\end{align} 

Thus, the regret only augments \textit{logarithmically} for the myopic rule, as
\begin{align}
    \mathcal{R}_t^\circ = \frac{\sqrt{\frac{32}{\pi}}\Delta\left|\mathrm{erf}^{-1}\left(2\Delta-1\right)\right| +\frac{4}{\sqrt{2\pi}}\int\limits_0^\infty  \Phi\left(-\sqrt{2}\left|\mathrm{erf}^{-1}\left(2\Delta-1\right)\right|-x/\sqrt{2}\right)\dd x}{ |\phi\left(\sqrt{2}\left|\mathrm{erf}^{-1}\left(2\Delta-1\right)\right|\right)|} \ln(t)+o(\ln(t)).
\end{align}
\begin{remark}
The $\sfrac{1}{t}$ decay of the threshold plays a key role in ensuring the sub-linear growth of the cumulative cost. One can show along identical lines that if a non-adaptive threshold is used, $p_t(\cdot)=\mathds{1}[|\cdot|\le \kappa]$ with $\kappa$ constant in time, one instead observes a \textit{linear} in time increase
\begin{align}
    \mathcal{R}_t=\left[2\Phi\left(\frac{\kappa}{\sqrt{q_0}}\right)-1\right]\Delta t +o(t),
\end{align}
no matter the margin $\kappa>0$.
\end{remark}

As an additional remark, note that the number of queried samples scales logarithmically with time, as the policy gets more stringent.
\begin{align}
\mathcal{Q}_t^\circ = \sqrt{\frac{2}{\pi}} \frac{4\left|\mathrm{erf}^{-1}\left(2\Delta-1\right)\right|}{\phi\left(\sqrt{2}\left|\mathrm{erf}^{-1}\left(2\Delta-1\right)\right|\right)}\ln(t) +o(\ln(t)).
\end{align}
To leading order,  
\begin{align}
    \varepsilon_{g,t}^\circ = \exp{- \frac{\phi\left(\sqrt{2}\left|\mathrm{erf}^{-1}\left(2\Delta-1\right)\right|\right)}{ \sqrt{\frac{2}{\pi}}4\left|\mathrm{erf}^{-1}\left(2\Delta-1\right)\right|}   \mathcal{Q}_t^\circ +o(\ln(t))}.
\end{align}
Namely, the myopic policy $\ploc_{\mathcal{R},t}(\cdot)$  makes the test error decay exponentially with the number of queries. This scaling is comparable to those proved for instance by \cite{dasgupta2005analysis}, for another selection policy. Finally, the cumulative number of mistakes also grows logarithmically as
\begin{align}
    \mathcal{M}^\circ_t = \frac{\frac{4}{\sqrt{2\pi}}\int\limits_0^\infty \dd x \,\Phi\left(-\sqrt{2}\left|\mathrm{erf}^{-1}\left(2\Delta-1\right)\right|-x/\sqrt{2}\right)}{ |\phi\left(\sqrt{2}\left|\mathrm{erf}^{-1}\left(2\Delta-1\right)\right|\right)|}\ln t+o(\ln(t)),
\end{align}
which concludes the derivation of Remark \ref{rem:scalings_locopt_QM}.

\subsection{Two-layer neural network}
We now turn to the derivation of Result \ref{res:2LNN_popt}, which borrows identical steps to that of Result \ref{res:loc_regret}.
At a given step $t$, conditional on the network weights $v,W$ at that time (with the latter assumed full rank), the average increment in cost reads
\begin{align}
    \frac{\dd\mathcal{R}_t}{\dd t}=\Delta \Ea{p_t(\omega)}+\Ea{(1-p_t(\omega))\mathds{1}[v^\top g(\Omega)\omega^\ast<0}],
\end{align}
where the expectation bears over 
\begin{align}
   \begin{pmatrix}
   \\
       \Omega\\
       \\
       \hline
       \omega^\ast
   \end{pmatrix} \sim \mathcal{N}\left(0_{r+1}, \begin{bmatrix}
        Q & \vline & M\\ \hline
        M^\top &\vline & 1
    \end{bmatrix}\right),
\end{align}
with $Q=\sfrac{WW^\top }{d}, M=\sfrac{W\beta}{d}$.
For a given $\Omega\in \R^r$, taking a functional derivative with respect to $p_t(\Omega)$, 
\begin{align}
    \frac{\delta}{\delta p_t(\Omega)}\frac{\dd\mathcal{R}_t}{\dd t}&=\frac{e^{-\frac{\Omega^\top Q^{-1} \Omega}{2}}}{\sqrt{(2\pi)^r \det{Q} }}\left[\Delta -\Ea{\mathds{1}[v^\top g(\Omega)\omega^\ast<0]\mid \Omega}\right]\\
    & =\frac{e^{-\frac{\Omega^\top Q^{-1} \Omega}{2}}}{\sqrt{(2\pi)^r \det{Q} }}\left[
    \Delta - \Phi\left(
    \frac{-\mathrm{sign}(v^\top g(\Omega)) M^\top Q^{-1} \Omega }{\sqrt{1-M^\top Q^{-1}M}}
    \right)
    \right].
\end{align}
We note that the assumption $\beta\notin {\rm rowsp}(W)$ ensures $1-M^\top Q^{-1}M>0$.
Thus, 

    \begin{align}
    \ploc_t(\Omega)=\mathds{1}\left[
   \Phi\left(
    \frac{-\mathrm{sign}(v^\top g(\Omega)) M^\top Q^{-1} \Omega }{\sqrt{1-M^\top Q^{-1}M}}
    \right)\ge \Delta
    \right],
\end{align}
which recovers \eqref{eq:opt_p_2LNN} of Result \ref{res:2LNN_popt}.

\section{Experimental details}\label{app:numerics}

\begin{figure}
    \centering
    \includegraphics[width=0.4\linewidth]{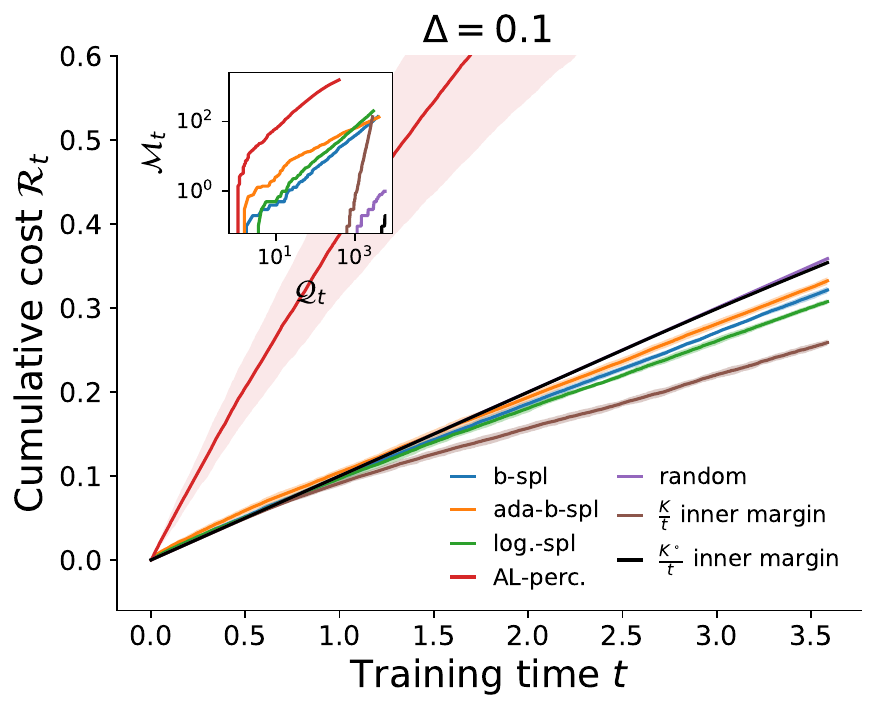}
    \includegraphics[width=0.4\linewidth]{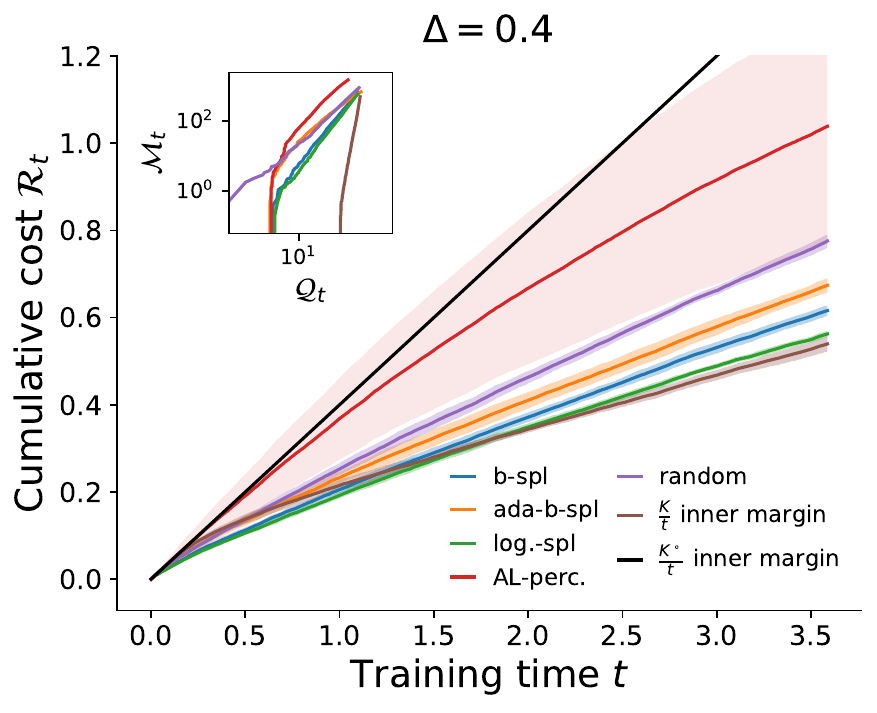}

    \caption{Same as Fig.\,\ref{fig:Benchmark_Gaussian} (right), for $\Delta=0.1, \,0.4$.}
    \label{fig:Benchmark_Chest}

\end{figure}

\begin{figure}
    \centering
    \includegraphics[width=0.4\linewidth]{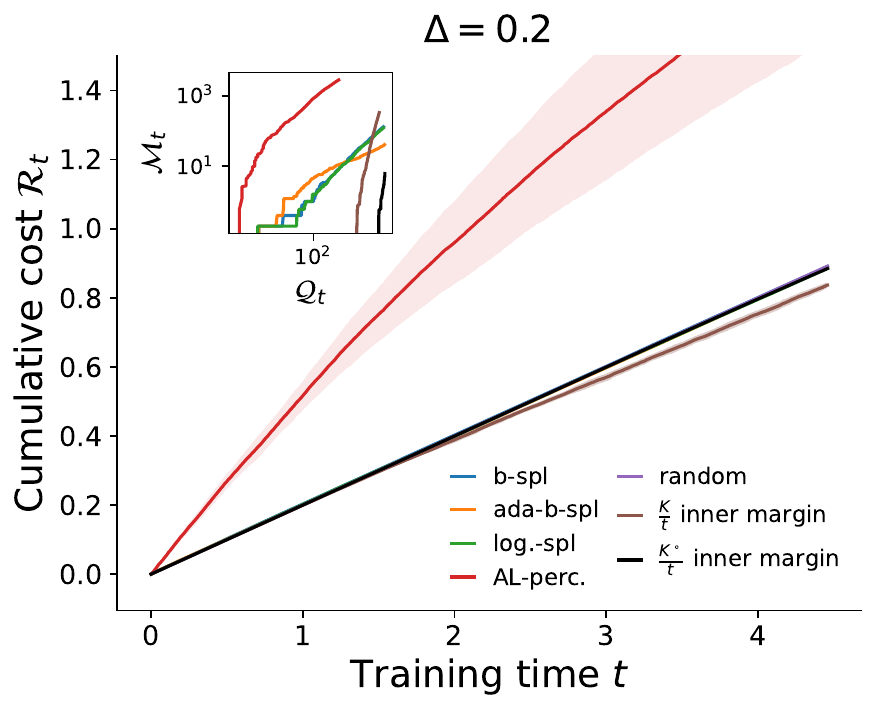}
    \includegraphics[width=0.4\linewidth]{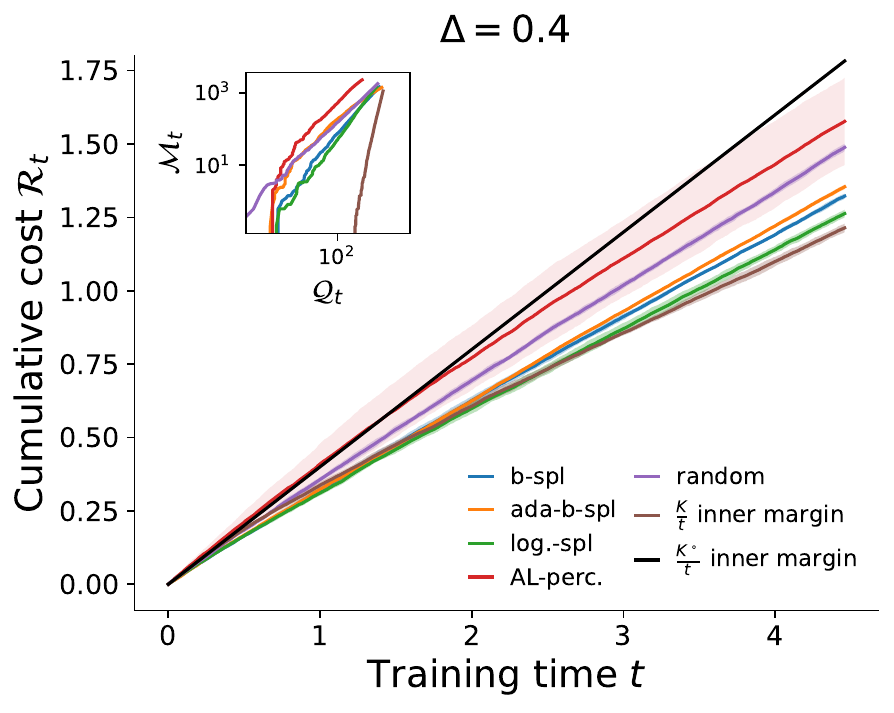}
    \caption{Same as Fig.\,\ref{fig:Benchmark_Gaussian}, for the Covid-19 diagnosis task from chest X-ray scans \cite{chowdhury2020can, rahman2021exploring}, for $\Delta=0.2, \,0.4$.}
    \label{fig:otherDelta_Covid}

\end{figure}

\begin{figure}
    \centering
    \includegraphics[width=0.4\linewidth]{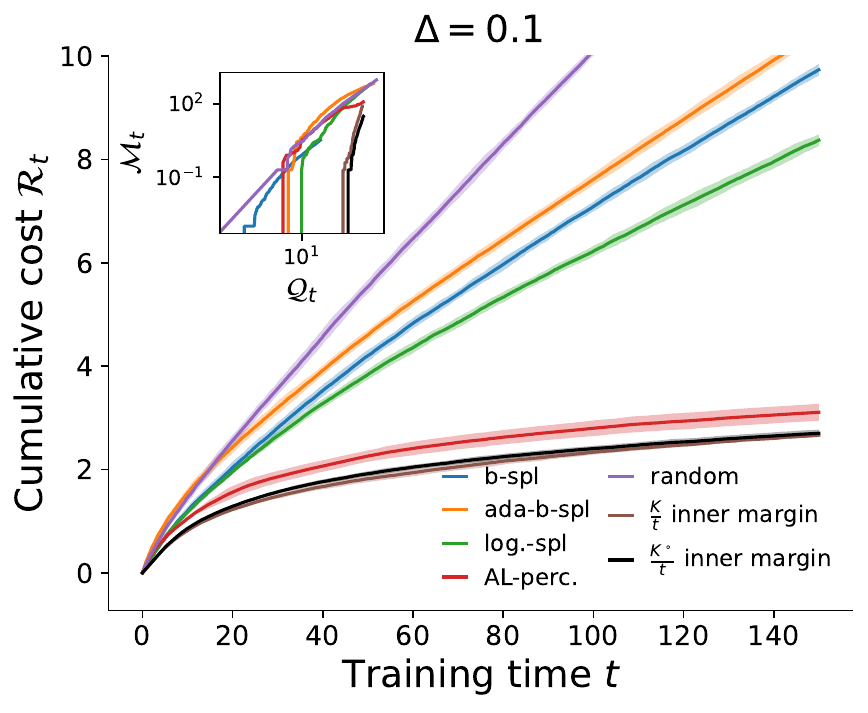}
 \includegraphics[width=0.4\linewidth]{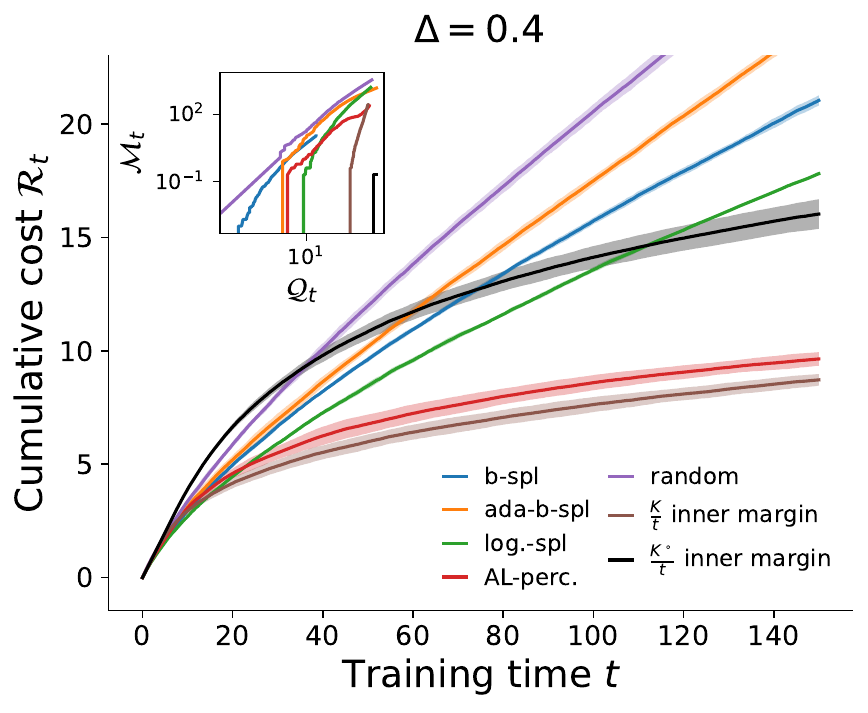}
    \caption{Same as Fig.\,\ref{fig:Benchmark_Gaussian} (left), for $\Delta=0.1, \,0.4$.}
    \label{fig:otherDelta_Gauss}

\end{figure}

\subsection{Benchmarked strategies}

We provide additional details on the implementation of the selection policies represented in Figs.\,\ref{fig:Benchmark_Gaussian}, \ref{fig:Benchmark_Chest}, which have been proposed in the literature for the problem considered in the present work, namely online active binary classification with a linear classifier trained with Adatron/Perceptron updates \cite{anlauf1989adatron, rosenblatt1958perceptron}. Let us note that all these algorithms exhibit comparable $O(dt_F)$ runtime.

\paragraph{b-sampling}(\texttt{b-spl}) \citep{cesa2004worst} proposes a margin-based selection rule with (constant-in-time) acquisition probability 
\begin{align}
    p_t(\cdot)=\frac{b}{b+|\cdot|},
\end{align}
for a given choice of bandwidth $b>0$. As specified in the original work, the Perceptron rule was used to update the model weights, instead of the Adatron rule.  For each one of Figs.\,\ref{fig:Benchmark_Gaussian}, \ref{fig:Benchmark_Chest}, \ref{fig:otherDelta_Covid}, \ref{fig:otherDelta_Gauss} a grid-search over $N=50$ values for the parameter $b$, evenly spread on a logarithmic grid between $10^{-2}, 10$, was performed, and only the trajectory corresponding to the value minimizing the terminal cumulative cost $\mathcal{R}_{t_F}$ was plotted.

\paragraph{Adaptive b-sampling}(\texttt{ada-b-spl}) \citep{cesa2004worst} is an adaptive version of b-sampling, which prescribes to increase the bandwidth $b$ as the square-root of the cumulative number of mistakes on queried samples, namely
\begin{align}
    p_t(\cdot)=\frac{\beta \sqrt{1+\tilde{\mathcal{M}}_t}}{\beta \sqrt{1+\tilde{\mathcal{M}}_t}+|\cdot|},
\end{align}
with 
\begin{align}
\tilde{\mathcal{M}}_t=\sum\limits_{i=1}^{td} \sigma_i\mathds{1}[\inprod{w_i, x_i}y_i<0]\,, 
\end{align}

for a given choice of tunable hyperparameter $\beta >0$. The value leading to minimal $\mathcal{R}_{t_F}$ over a logarithmic grid of size $50$ between $10^{-2}, 10$ was plotted.

\paragraph{Logistic sampling} (\texttt{log.-spl}) \citep{sculley2007online} instead proposes a constant-in-time \textit{exponential} (rather than sigmoid) ansatz for the acquisition probability
\begin{align}
    p_t(\cdot)=e^{-\gamma |\cdot|},
\end{align}
with hyperparameter $\gamma>0$, again leveraging the Perceptron training rule. The value leading to minimal $\mathcal{R}_{t_F}$ over a logarithmic grid of size $50$ between $10^{-2}, 10$ was plotted.

\paragraph{Active Perceptron} (\texttt{AL-Perc}) \citep{dasgupta2005analysis} corresponds to an Adatron rule with learning rate $\eta=2$, and a hard inner-margin threshold policy
\begin{align}
    p_t(\cdot)=\mathds{1}[|\cdot|\le s_t],
\end{align}
where the threshold $s_t$ is initialized at $\sfrac{1}{\sqrt{d}}$, and divided by $2$ every time a streak of $R \in \mathbb{N}$ consecutive correct predictions is achieved. It thus resembles the myopic policy \eqref{eq:scaling_kappat} of Result \ref{res:loc_regret}, in that it involves a shrinking margin. $50$ values of the patience parameter $R$ between $1$ and $ 101$ were simulated, and the value leading to minimal terminal cumulative cost plotted. Note that the pseudo-code of the original paper \cite{dasgupta2005analysis} does not explicitly prescribe to reset the streak count after every threshold-halving, but the subsequent work of \cite{monteleoni2007practical} does. We implement the latter prescription.

\paragraph{Passive baseline} (\texttt{random}) We finally include the baseline of a random selection policy, which queries a sample with constant prediction $p\in [0,1]$, irrespective of the sample. $50$ values of $p \in[0.01, 1]$ were simulated, and the value minimizing the terminal cost retained and plotted. Updates are performed using the Perceptron rule.

In addition to these classical baselines, we implement the following policies, taking inspiration from the characterization of the myopic policy of Res.\,\ref{res:loc_regret}.

\paragraph{Inner-margin with shrinking threshold} ($\sfrac{K}{t}$-\texttt{inner-margin}) corresponds to a inner-margin policy 
\begin{align}
    p_t(\cdot)=\mathds{1}[|\cdot|\le \sfrac{K}{t}],
\end{align}
with tunable hyperparameter $K$. In  Figs.\,\ref{fig:Benchmark_Gaussian}, \ref{fig:Benchmark_Chest}, \ref{fig:otherDelta_Covid}, \ref{fig:otherDelta_Gauss}, a grid-search over $N=50$ values for the parameter $K$, evenly spread on a logarithmic grid between $10^{-2}, 50$, was performed, and only the trajectory corresponding in hindsight to the value minimizing the terminal cumulative cost $\mathcal{R}_{t_F}$ was plotted. 

\paragraph{Inner-margin with theory-prescribed threshold} ($\sfrac{K^\circ}{t}$-\texttt{inner-margin}) Note that Res.\,\ref{res:loc_regret}, Eq.\,\ref{eq:scaling_kappat} affords a particular prescription for the parameter $K$ above. The black lines in Figs.\,\ref{fig:Benchmark_Gaussian}, \ref{fig:Benchmark_Chest}, \ref{fig:otherDelta_Covid}, \ref{fig:otherDelta_Gauss} correspond to this particular value $K=K^\circ$. For instance, for $\Delta=0.2, q_0=0.1, \eta=2$, $K^\circ\approx 2.52$. In contrast to all other baselines, this policy is hyperparameter-free, and calls for no tuning.

We note that the inner-margin policy performs comparatively well against all studied baselines in all probed settings, provided the hyperparameter $K$ is tuned. The particular prescription $K^\circ$ \eqref{eq:expression_kappa} on the other hand performs poorly on real datasets, and is found to query too often.

\subsection{Preprocessing of the datasets}

\paragraph{Covid 19 dataset}\citep{chowdhury2020can, rahman2021exploring} The original dataset includes $3616$ chest X-ray scans from Covid-positive patients, and $17549$ scans from patients with no infection, non-Covid infection, pneumonia. For simplicity, the latter two classes were discarded, and the class of no infection was downsampled to $3616$ images to balance the dataset, yielding a total of $7232$ samples. Each image was resized to $40 \times 40$ pixel. A held-out dataset was constituted from $100$ samples, and used to fit the \texttt{scikit-learn} \citep{scikit-learn} \texttt{StandardScaler}. The latter was then used to transform the remaining $n=7132$ images, which corresponds to the total stream length. At each round, the sample is normalized to unit norm before being presented to the model and policy.

\paragraph{Chest-Xray scans dataset} \citep{kermany2018identifying} The dataset includes $5863$ chest X-ray scans from patients with(out) pneumonia. The two classes were not balanced. Each image was resized to $40 \times 40$ pixel. A held-out dataset was constituted from $100$ samples, and used to fit the \texttt{scikit-learn} \citep{scikit-learn} \texttt{StandardScaler}. The latter was then used to transform the remaining $n=5763$ images, which corresponds to the total stream length. At each round, the sample is normalized to unit norm before being presented to the model and policy.

\subsection{Training of the 2-layer neural network}

We now detail the training procedure followed to obtain the network weights $v,W$, from which Fig.\, \ref{fig:2LNN} was then produced, using the expression of Result \ref{res:2LNN_popt}. The readout weights $v$ were initialized, and kept fixed throughout training, to the all-ones vector $v=(1,1)^\top$. The weights $W$ were initialized with i.i.d standard Gaussian entries, then updated $n=2000$ times with the online gradient update
\begin{align}
    W_{i+1}=W_i-\eta \ell^\prime\left(v^\top g(W_ix_{i+1}), \sgn(\inprod{\beta, x_{i+1}})\right)(v \odot g^\prime(W_ix_{i+1}))x_{i+1}^\top 
\end{align}
with unit learning rate $\eta=1$, on the logistic loss $\ell(z,y)=\ln(1+\exp{-yz})$, and without any data selection. The activation $g(\cdot)$ was taken to be GeLU. Simulations were conducted in dimension $d=500$.

\section{Logistic loss}\label{app:other_losses}
The main text focuses on the Adatron online training procedure \citep{anlauf1989adatron, engel2001statistical}, which can be viewed as a variant of the classical Perceptron rule \citep{rosenblatt1958perceptron}. While these losses are classically considered in many works on online active classification \citep{dasgupta2004analysis, cesa2004worst, sculley2007online, cesa2009robust}, losses such as the logistic loss are more commonly leveraged in day-to-day machine learning practice. The present Appendix is devoted to expanding the discussion of the main text the latter, highlighting in particular the difference with Adatron/Perceptron rules. In fact, the use of the logistic loss in the SGD updates \eqref{eq:discrete_GD_step} instead of the Adatron/Perceptron loss leads to a quantitatively different behavior, notably in terms of growth rates of the cumulative cost. This is the object of the present Appendix.

\subsection{Myopic policy for the cumulative cost}

We revisit the analysis of the myopic rule $\ploc_{\mathcal{R},t}$ for the cumulative cost \eqref{eq:Rk}, characterized for the Adatron rule in Result \ref{res:loc_regret}, for the case of the logistic loss. While $\ploc_{\mathcal{R},t}$ still takes the form \eqref{eq:p_loc_opt_regret}, since the latter only depend on the loss through the current-time summary statistics values $m_t^\circ, q_t$, the asymptotic behavior of the threshold $\kappa^\circ_t$ differs. The following conjecture surmises that the latter tends to a non-zero limit in the large-time limit $t\to\infty$, as opposed to the $\sfrac{1}{t}$ decay for Adatron.

\begin{figure}
    \centering
    \includegraphics[width=0.5\linewidth]{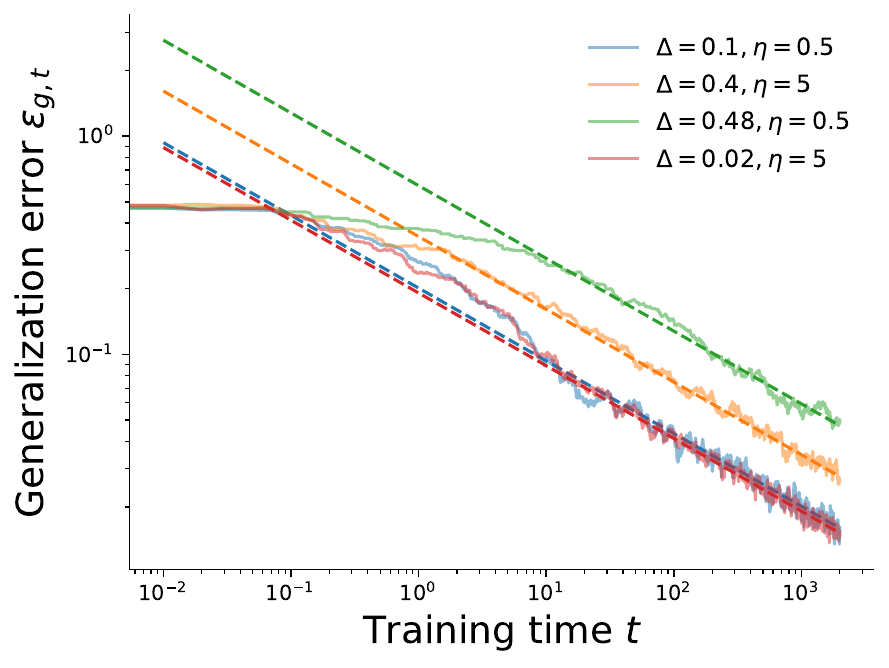}
    \caption{Test error $\epsilon_{g,t}$ as a function of training time, with online SGD with logistic loss, under the myopic policy $\ploc_{\mathcal{R},t}$ \eqref{eq:p_loc_opt_regret}. Solid lines represent single runs in dimension $d=100$, dashed lines represent the theoretical prediction for the $t\to\infty$  asymptotics of Conjecture \ref{conj:opt_other_losses}.}
    \label{fig:scaling_logistic}
\end{figure}

\begin{conjecture}[Myopic policy for the cumulative cost, logistic loss] \label{conj:opt_other_losses}
    Let $\ell$ be the logistic loss $\ell(z)=\ln(1+\exp{-z})$. If the equation
\begin{align} \label{eq:Necessary_condition}
   \int\limits_{u=0}^\infty  \int\limits_{v=\frac{-\sqrt{2}z\left|\mathrm{erf}^{-1}\left(2\Delta-1\right)\right|-u}{z}}^{\frac{\sqrt{2}z\left|\mathrm{erf}^{-1}\left(2\Delta-1\right)\right|-u}{z}} \frac{e^{-\frac{v^2}{2}}}{2\pi}\left[
   \ell^\prime(u+zv)\left(2zv-\eta \ell^\prime(u+zv))\right)
   \right] \dd u \dd v =0
\end{align}
admits a positive solution $z^\ast >0$, then the threshold $\kappa^\circ_t$ \eqref{eq:expression_kappa} admits a non-zero limit
\begin{align}
 \kappa^\circ_t \xrightarrow[]{t\to \infty}\sqrt{2}z^\ast\left|\mathrm{erf}^{-1}\left(2\Delta-1\right)\right|.
\end{align}
Furthermore, when the policy $\ploc_{\mathcal{R},t}$ is employed, the test error decays asymptotically as
\begin{align}
    \varepsilon_{g,t}^\circ = \frac{\sfrac{z^\ast}{\pi}}{\left(-3 \eta\left[\int\limits_{u=0}^\infty  \int\limits_{v=\frac{-\sqrt{2}z^\ast\left|\mathrm{erf}^{-1}\left(2\Delta-1\right)\right|-u}{z^\ast}}^{\frac{\sqrt{2}z^\ast\left|\mathrm{erf}^{-1}\left(2\Delta-1\right)\right|-u}{z^\ast}} \frac{e^{-\frac{v^2}{2}}}{2\pi}\left[
   \ell^\prime(u+z^\ast v)\left(2(u+z^\ast v)-\eta \ell^\prime(u+z^\ast v))\right)
   \right]\dd u\dd v\right] \right)^{\frac{1}{3}}} t^{-\frac{1}{3}}+o( t^{-\frac{1}{3}})
\end{align}
\end{conjecture}
As will be detailed in the following subsection, Eq.\,\eqref{eq:Necessary_condition} is in fact a necessary condition for the following two asymptotic statements to be self-consistent. Heuristically, we numerically observe the predicted scaling and leading asymptotic term across a broad range of values for $\Delta,\eta$ (see Fig.\,\ref{fig:scaling_logistic}), leading us to surmise that Conj. \ref{conj:opt_other_losses} holds more broadly.

An immediate corollary of Conj. \ref{conj:opt_other_losses} is that although the cumulative cost \eqref{eq:Rk}, number of mistakes \eqref{eq:Mk} and number of queries \eqref{eq:Qk} still admit sub-linear growth, the latter is no longer logarithmic, but power-law.

\begin{remark}[Cumulative cost scaling]
    The cumulative cost $\mathcal{R}^\circ_t$, number of mistakes $\mathcal{M}^\circ_t$, and number of queries $\mathcal{Q}^\circ_t$ scale as $\mathcal{R}^\circ_t,\mathcal{M}^\circ_t,\mathcal{Q}^\circ_t\asymp t^{\sfrac{2}{3}}$.
\end{remark}
This power-law is to be contrasted to the much milder logarithmic growth induced by the Adatron loss, as summarized in Rem. \ref{rem:scalings_locopt_QM}. 

\subsection{Derivation of Conj. \ref{conj:opt_other_losses} }

In order to derive the asymptotic scaling of $\kappa^\circ_t$ \eqref{eq:expression_kappa}, one needs to study the dynamics of the summary statistics $m_t^\circ, q_t^\circ$, as governed by the ODEs \eqref{eq:ODEs} controlled by the myopic policy $\ploc_{\mathcal{R},t}$ \eqref{eq:p_loc_opt_regret}. In the likeness of App. \ref{app:Localopt_R}, we make a series of ansatz, which will be subsequently verified in self-consistent fashion. We assume
\begin{align}
    m_t^\circ, q_t^\circ \xrightarrow[]{t\to\infty} \infty,
\end{align}
together with
\begin{align}
 \gamma_t^\circ \xrightarrow[]{t\to\infty} 1, && \sqrt{2(1-\gamma_t^\circ)q_t^\circ} \xrightarrow[]{t\to\infty} z,
\end{align}
where we introduced the limit $z>0$, to be determined self-consistently. Under these assumptions, the threshold $\kappa^\circ_t$ then converges to the limit
\begin{align}
 \kappa^\circ_t \xrightarrow[]{t\to \infty}\sqrt{2}z\left|\mathrm{erf}^{-1}\left(2\Delta-1\right)\right|.
\end{align}
It thus remains to ascertain $z$. On the one hand, the ODE for $m_t^\circ$ simplifies as
 \begin{align}
     \frac{\dd m_t^\circ}{\dd t}&=-2\eta \int\limits_0^\infty \frac{e^{-\frac{1}{2}(\omega^\ast)^2}}{\sqrt{2\pi }} \int\limits_{\frac{-\sfrac{\kappa^\circ_t}{\sqrt{q_t^\circ}}-\omega^\ast}{\sqrt{2(1-\gamma)}}}^{\frac{\sfrac{\kappa^\circ_t}{\sqrt{q_t^\circ}}-\omega^\ast}{\sqrt{2q_t^\circ(1-\gamma_t^\circ)}}}\frac{e^{-\frac{1}{2}g^2}}{\sqrt{2\pi }}\ell^\prime(m_t^\circ \omega^\ast +\sqrt{2q_t^\circ(1-\gamma_t^\circ)}g)\omega^\ast \dd \omega^\ast \dd g\\
     &\approx -2\frac{\eta}{q_t^\circ} \int\limits_0^\infty \frac{e^{-\frac{1}{2q_t^\circ}(\omega^\ast)^2}}{\sqrt{2\pi }} \int\limits_{\frac{-\kappa^\circ_t-\omega^\ast}{\sqrt{2q_t^\circ(1-\gamma_t^\circ)}}}^{\frac{\kappa^\circ_t-\omega^\ast}{\sqrt{2q(1-\gamma_t^\circ)}}}\frac{e^{-\frac{1}{2}g^2}}{\sqrt{2\pi }}\ell^\prime(\omega^\ast +\sqrt{2q_t^\circ(1-\gamma_t^\circ)}g)\omega^\ast\dd \omega^\ast \dd g\\
     &\approx -\frac{2\eta}{q}   \int\limits_0^\infty \int\limits_{\frac{-\sqrt{2}z\left|\mathrm{erf}^{-1}\left(2\Delta-1\right)\right|-u}{z}}^{\frac{\sqrt{2}z\left|\mathrm{erf}^{-1}\left(2\Delta-1\right)\right|-u}{z}} \frac{e^{-\frac{v^2}{2}}}{2\pi}\left[
   \ell^\prime(u+zv)u
   \right]\dd u \dd v.
 \end{align}
By the same token
 \begin{align}
 \frac{\dd q_t^\circ}{\dd t}\approx -\frac{2\eta}{\sqrt{q}}\left[ \int\limits_0^\infty \int\limits_{\frac{-\sqrt{2}z\left|\mathrm{erf}^{-1}\left(2\Delta-1\right)\right|-u}{z}}^{\frac{\sqrt{2}z\left|\mathrm{erf}^{-1}\left(2\Delta-1\right)\right|-u}{z}} \frac{e^{-\frac{v^2}{2}}}{2\pi}\left[
   \ell^\prime(u+z v)\left(2(u+z v)-\eta \ell^\prime(u+z v))\right)
   \right] \dd u \dd v\right]
 \end{align}
 Thus, 
\begin{align}
    &m_t^\circ= \left(-6 \eta   \int\limits_0^\infty \int\limits_{\frac{-\sqrt{2}z\left|\mathrm{erf}^{-1}\left(2\Delta-1\right)\right|-u}{z}}^{\frac{\sqrt{2}z\left|\mathrm{erf}^{-1}\left(2\Delta-1\right)\right|-u}{z}} \frac{e^{-\frac{v^2}{2}}}{2\pi}\left[
   \ell^\prime(u+zv)u
   \right]\dd u \dd v\right)^{\frac{1}{3}} t^{\frac{1}{3}}+o( t^{\frac{1}{3}}),\\
    & q_t^\circ\asymp \left(- 3 \eta\left[ \int\limits_0^\infty \int\limits_{\frac{-\sqrt{2}z\left|\mathrm{erf}^{-1}\left(2\Delta-1\right)\right|-u}{z}}^{\frac{\sqrt{2}z\left|\mathrm{erf}^{-1}\left(2\Delta-1\right)\right|-u}{z}} \frac{e^{-\frac{v^2}{2}}}{2\pi}\left[
   \ell^\prime(u+z v)\left(2(u+z v)-\eta \ell^\prime(u+z v))\right)
   \right]\dd u \dd v \right]  \right)^{\frac{2}{3}}t^{\frac{2}{3}}+o( t^{\frac{2}{3}}).
\end{align}
From the above and the observation that $\varepsilon_{g,t}^\circ\asymp \sfrac{\sqrt{1-\gamma} }{\pi}$, one deduces
\begin{align}
    \varepsilon_{g,t}^\circ \asymp \frac{\sfrac{z}{\pi}}{\left(- 3 \eta\left[ \int\limits_0^\infty \int\limits_{\frac{-\sqrt{2}z\left|\mathrm{erf}^{-1}\left(2\Delta-1\right)\right|-u}{z}}^{\frac{\sqrt{2}z\left|\mathrm{erf}^{-1}\left(2\Delta-1\right)\right|-u}{z}} \frac{e^{-\frac{v^2}{2}}}{2\pi}\left[
   \ell^\prime(u+z v)\left(2(u+z v)-\eta \ell^\prime(u+z v))\right)
   \right] \dd u \dd v\right] \right)^{\frac{1}{3}}} t^{-\frac{1}{3}}.
\end{align}
One is now in a position to determine self-consistently $z$. For all the above to be consistent, one has, on the one hand
\begin{align}
\label{eq:dgdt1}
    \frac{\dd \gamma_t^\circ}{\dd t}= -\frac{\dd}{\dd t}\left(\frac{z^2}{2q_t^\circ}\right)=\frac{z^2}{2(q_t^\circ)^2} \frac{\dd q_t^\circ}{\dd t} \asymp t^{-\frac{5}{3}}.
\end{align}

On the other hand, from the definition of $\gamma_t^\circ$, one also has

\begin{align}\label{eq:dgdt2}
   \frac{\dd  \gamma_t^\circ}{\dd t}= &- \eta \frac{2}{(q_t^\circ)^{3/2}}\int\limits_0^\infty \int\limits_{\frac{-\sqrt{2}z\left|\mathrm{erf}^{-1}\left(2\Delta-1\right)\right|-u}{z}}^{\frac{\sqrt{2}z\left|\mathrm{erf}^{-1}\left(2\Delta-1\right)\right|-u}{z}} \frac{e^{-\frac{v^2}{2}}}{2\pi}\left[
   \ell^\prime(u+zv)u
   \right] \dd u\dd v\\
   &+\frac{m_t^\circ \eta}{(q_t^\circ)^{2}}\left[ \int\limits_0^\infty \int\limits_{\frac{-\sqrt{2}z\left|\mathrm{erf}^{-1}\left(2\Delta-1\right)\right|-u}{z}}^{\frac{\sqrt{2}z\left|\mathrm{erf}^{-1}\left(2\Delta-1\right)\right|-u}{z}} \frac{e^{-\frac{v^2}{2}}}{2\pi}\left[
   \ell^\prime(u+z v)\left(2(u+z v)-\eta \ell^\prime(u+z v))\right)
   \right]\dd u\dd v\right]\\
   & \approx \frac{\eta}{(q_t^\circ)^{\sfrac{3}{2}}} \int\limits_{u=0}^\infty  \int\limits_{v=\frac{-\sqrt{2}z\left|\mathrm{erf}^{-1}\left(2\Delta-1\right)\right|-u}{z}}^{\frac{\sqrt{2}z\left|\mathrm{erf}^{-1}\left(2\Delta-1\right)\right|-u}{z}} \frac{e^{-\frac{v^2}{2}}}{2\pi}\left[
   \ell^\prime(u+zv)\left(2zv-\eta \ell^\prime(u+zv))\right)
   \right] \dd u\dd v+o((q_t^\circ)^{-\sfrac{3}{2}})
\end{align}
Unless the double integral is zero, this scales as $(q_t^\circ)^{-\sfrac{3}{2}}\asymp t^{-1}$. Since equations \eqref{eq:dgdt1} and \eqref{eq:dgdt2} are equal, the right hand side of the latter must vanish at leading $ t^{-1}$ order. This provides the following condition 
\begin{align}\label{eq:selfconsz}
\int\limits_{u=0}^\infty  \int\limits_{v=\frac{-\sqrt{2}z\left|\mathrm{erf}^{-1}\left(2\Delta-1\right)\right|-u}{z}}^{\frac{\sqrt{2}z\left|\mathrm{erf}^{-1}\left(2\Delta-1\right)\right|-u}{z}} \frac{e^{-\frac{v^2}{2}}}{2\pi}\left[
   \ell^\prime(u+zv)\left(2zv-\eta \ell^\prime(u+zv))\right)
   \right] \dd u\dd v=0.
\end{align}
We have thus established that if there exists a regime where the above assumed convergences are realized, then Eq. \eqref{eq:selfconsz} necessarily admits a positive solution. We numerically observe that the quantitative asymptotics of Conj. \ref{conj:opt_other_losses} hold for a broad section of values for $\Delta, \eta$.

\end{document}